\documentclass[a4paper,fleqn]{cas-dc}

\usepackage[authoryear,longnamesfirst]{natbib}
\usepackage{pifont}
\usepackage{multirow}
\usepackage{bm}
\usepackage{makecell}
\usepackage{amsmath}

\def\tsc#1{\csdef{#1}{\textsc{\lowercase{#1}}\xspace}}
\tsc{WGM}
\tsc{QE}

\begin{document}
\let\WriteBookmarks\relax
\def\floatpagepagefraction{1}
\def\textpagefraction{.001}

\shorttitle{Bidirectional Face-Bone Transformation via TCFNet}

\title [mode = title]{TCFNet: Bidirectional Face-Bone Transformation via a Transformer-Based Coarse-to-Fine Point Movement Network}

\tnotemark[1] 

\tnotetext[1]{This work was supported by the National Key Research and Development Program of China (Grant No. 2022YFC2405401),  the Natural Science Foundation of Beijing Municipality under Grant L232037, the Natural Science Foundation of China (Grant No. 62173014 and Grant No. U22A2051). (Corresponding author: Junchen Wang.)} 

%

\author[1]{Runshi Zhang}[
]

\affiliation[1]{organization={School of Mechanical Engineering and Automation, Beihang University},
	addressline={37 Xueyuan Road, Haidian District}, 
	postcode={10191}, 
	state={Beijing},
	country={China}}

\author[2]{Bimeng Jie}[
]
\author[2]{Yang He}[
]
\affiliation[2]{organization={Peking University School and Hospital of Stomatology},
	addressline={Weigong Village, Haidian District}, 
	postcode={100081}, 
	state={Beijing},
	country={China}}

\author[1]{Junchen Wang}[
orcid=0000-0002-0916-9932]

\cormark[1]

\ead{wangjunchen@buaa.edu.cn}

\ead[url]{https://mrs.buaa.edu.cn/}
\begin{abstract}
Computer-aided surgical simulation is a critical component of orthognathic surgical planning, where accurately simulating face-bone shape transformations is significant.
The traditional biomechanical simulation methods are limited by their computational time consumption levels, labor-intensive data processing strategies and low accuracy.
Recently, deep learning-based simulation methods have been proposed to view this problem as a point-to-point transformation between skeletal and facial point clouds.
However, these approaches cannot process large-scale points, have limited receptive fields that lead to noisy points, and employ complex preprocessing and postprocessing operations based on registration.
These shortcomings limit the performance and widespread applicability of such methods.
Therefore, we propose a Transformer-based coarse-to-fine point movement network (TCFNet) to learn unique, complicated correspondences at the patch and point levels for dense face-bone point cloud transformations.
This end-to-end framework adopts a Transformer-based network and a local information aggregation network (LIA-Net) in the first and second stages, respectively, which reinforce each other to generate precise point movement paths.
LIA-Net can effectively compensate for the neighborhood precision loss of the Transformer-based network by modeling local geometric structures (edges, orientations and relative position features).
The previous global features are employed to guide the local displacement using a gated recurrent unit.
Inspired by deformable medical image registration, we propose an auxiliary loss that can utilize expert knowledge for reconstructing critical organs.
Our framework is an unsupervised algorithm, and this loss is optional.
Compared with the existing state-of-the-art (SOTA) methods on gathered datasets, TCFNet achieves outstanding evaluation metrics and visualization results.
The code is available at https://github.com/Runshi-Zhang/TCFNet.
\end{abstract}

\begin{highlights}
\item We propose a bidirectional point cloud transformation framework to simulate the relationship between skeletal and facial appearances during surgical planning for orthognathic surgery.
\item To overcome the neighborhood precision loss of the Transformer-based module, we present a local information aggregation network for modelling local geometric structures and enhancing local feature representations.
\item Inspired by the auxiliary loss used for deformable medical image registration, an optional auxiliary loss is proposed to facilitate correspondence between local critical structures.
\item Compared with other SOTA methods, our approach achieves outstanding results in terms of evaluation metrics and visualizations.
\end{highlights}


\maketitle

\section{Introduction}
Orthognathic surgery can correct craniomaxillofacial (CMF) deformities, restore facial appearances and enhance bone functions, such as chewing and swallowing~\cite{Zammit}. 
Its primary goal is to correct skeletal deformities and establish a harmonized spatial relationship between the maxilla and mandible~\cite{millesi2023surgery}. 
Additionally, facial aesthetic harmonization, a major concern for patients, is an important goal of orthognathic surgery~\cite{bianchi2023beauty}.
Orthognathic surgery often relies on computer-assisted surgery (CAS)~\cite{Mohammed} for preoperative planning, which determines the displacement for normalizing the skeletal deformities of osteotomized bones by modeling the maxilla and mandible.
Specifically, the process includes (1) segmenting and reconstructing the maxilla and mandible from (cone beam) computed tomography (CBCT, CT) slices~\cite{zhangrunshi};
(2) extracting the osteotomized bones and simulating the surgical procedure to determine the optimal movement path~\cite{rhan,xifang,maleitmi}, and
(3) adjusting the osteotomized bones to normalize them and ensure that the facial surface changes align with improved deformity correction.
Therefore, precise preoperative planning during CAS plays a crucial role in the success of orthognathic surgery and the postoperative recovery of the patient~\cite{RafaelDenadai}.
Computer-aided surgical simulation (CASS) technology can accurately simulate the adjustment paths of osteotomized bones.
Although skeletal movements can postoperatively achieve excellent functional occlusion, achieving an aesthetically harmonious, proportional, and pleasing facial appearance is challenging due to the complex and nonlinear interactions between bones and soft tissue. 
Thus, accurately simulating the relationships between bones and facial appearances is critically important in the planning task of orthognathic surgery.

Face-bone transformation aims to reconstruct the appearance of a face from its skeletal structure and vice versa; biomechanical simulation methods are widely employed for prediction~\cite{MollemansW}.
These methods usually formulate handcrafted features to predict targeted tissues and include the mass spring model (MSM), mass tensor model (MTM) and finite-element model (FEM)~\cite{KimDaeseung_miccai}. Among them, the FEM often achieves the best prediction accuracy~\cite{RuggieroFederica}. 
However, the FEM requires significant computational time and a labor-intensive data processing scheme to produce acceptable FE mesh models in a clinical pipeline~\cite{KimDaeseung_mia}. 
Its practicality is limited since surgeons often need to make multiple attempts during orthognathic surgical planning processes, whereas simulations must be conducted quickly to be effective in clinical pipelines~\cite{LampenNathan}.
In addition, most methods assume that similar tissues maintain consistency after they transform, which is unrealistic as this assumption overlooks the subject-specific variability aspect.

Recent advancements in deep learning, particularly in 3D representation learning methods for point cloud processing, offer promising alternatives to the traditional biomechanical simulation methods for efficiently and swiftly providing tissue simulation predictions~\cite{LampenNathan,xifang,maleitmi,malei,BaoJ}.
Deep learning-based methods approach face-bone transformation as a point-to-point correspondence problem between pairs of point clouds. 
Specifically, dense point clouds are sampled from facial and skeletal meshes to ensure accurate reconstruction results without distortions. 
A point cloud learning method is then employed to extract detailed topological and structural relationships for transformation purposes, resulting in a point-to-point displacement field. 
This field is used to move points from the source 3D shape to the target shape.
Previous works~\cite{malei,BaoJ} utilized 3D discrete convolutions to extract features from a voxelized 3D space for use in deep neural networks implemented on point clouds; this strategy induces many outliers and incurs high levels of computational and memory consumption~\cite{WuWenxuan}.
Therefore, these implementations are memory consuming and inefficient, limiting the network input to fewer than 5000 points. 
This constraint makes it challenging for previous methods to leverage the potential benefits of large-scale data, rendering them unsuitable for handling dense point clouds. 
It is obvious that processing sparse point clouds can reduce the capability of modeling local detail features and depends on the performance of the point cloud sampling algorithm. 
The previous networks are able to enlarge receptive field with using downsampling layer, but the receptive field is limited compared to Transformer-based methods.

In response to large-scale point clouds, recent advancements~\cite{GrahamBenjamin} have made progress in terms of producing efficient convolutional backbones that can overcome the above data scale limitation.  
Sparse convolution~\cite{GrahamBenjamin} alleviates this limitation, but convolutional networks with small receptive fields struggle to capture global dependencies.
Transformer-based models~\cite{ZhaoHengshuang} are particularly suitable for point cloud processing since the self-attention mechanism, the core of the Transformer architecture, is essentially a set operator and exhibits input permutation invariance.
A recent work~\cite{WuXiaoyangv3} introduced a Transformer-based model with stronger performance named Point Transformer V3 (PTV3), a wider receptive field and a faster speed.
Specifically, it replaces the traditional K-nearest neighbors (KNN) operation with point cloud serialization, allowing for more efficiently handling of dense point clouds. 
Additionally, it uses Transformer to capture spatial global relationships and expands the receptive field to 1024 points.

Therefore, we propose a Transformer network based on Point Transformer V3 (PTv3)~\cite{WuXiaoyangv3} to efficiently process dense face-bone point clouds.
However, the performance achieved when modeling local structures in PTv3 degrades because the K-nearest neighbors (KNN) operation is discarded.
To overcome this problem and inspired by the 3D shape completion community, which widely uses multistep or multiscale frameworks~\cite{PMPNetplus, PMPNet, YuXumin} to recover missing point clouds in a coarse-to-fine manner, we introduce TCFNet, which is a Transformer-based coarse-to-fine point movement network.
Specifically, TCFNet adopts the following adaptations to achieve outstanding accuracy and efficient performance in face-bone shape transformation tasks.

(1) We propose a bidirectional point cloud transformation framework named TCFNet for bidirectional face-bone prediction.
It adopts a two-step coarse-to-fine network that includes a Transformer-based stage and a local information aggregation network (LIA-Net) to learn the dense correspondences and complex local structures between facial and skeletal point clouds.
Compared with the previously developed methods~\cite{malei}, TCFNet can achieve face-bone shape transformation in a single test without downsampling into sparse point clouds, and several of its evaluation metrics and visualization results are superior.
Moreover, it does not require prealignment during preprocessing or template registration during postprocessing.

(2) To overcome the neighborhood precision loss induced by point cloud serialization, we propose LIA-Net to model local geometric structures that combine local region features, their orientations, and relative global position features.
To aggregate the features of the previous moving path and its current location, the global features in the first stage are employed to guide the local displacement field in LIA-Net, akin to a gated recurrent unit (GRU). 
Ablation experiments show that our proposed LIA-Net model significantly enhances the ability to transform and reconstruct local structures, such as the nose and lip.

(3) Inspired by the auxiliary loss used in deformable medical image registration tasks, we present an optional auxiliary loss to facilitate correspondences between the local critical features of organs that are of great concern to doctors.
Similar to registration, massive organ annotations can improve the local and global performance of the transformation and reconstruction processes.
\section{Related work}
\subsection{Face-Bone Modeling}
Face-bone modeling is a long-standing computer vision (CV) and medical surgical planning problem.
Early face-bone modeling methods were proposed to generate mapping relationships via statistical shape models~\cite{PaysanPascal,BinJia,ZhangDan}, and these approaches are widely used in archaeology, anthropology and forensic science.
These statistical models are dependent on the assumption that similar source shapes must necessarily be similar to the target shape after the transformation process, which is unrealistic.
In the CV community, the existing 3D facial reconstruction methods can be categorized according to their 2D inputs as multiview, monocular video, single-image reconstruction and sketch-based modeling techniques~\cite{RMallikarjunB,XiaoDeqiang,WangLizhen}.
These methods rely on highly precise images or videos, which is challenging due to the limited information and the high level of occlusion contained in photos of daily life ~\cite{XiaoDeqiang}.
In addition, these methods fall short in terms of representing 3D facial models with customized surface details, resulting in an inability to directly adopt them in surgical planning tasks~\cite{LuoZhongjin}.
In the medical image analysis community, D. Xiao~\cite{XiaoDeqiang,XiaoDeqianghuiyi} proposed the geometric deformation method to deform a mean face with an estimated deformation field and then obtain the 3D facial surface of the patient.
The predicted bone was retrieved from a normal facial and bony shape database.
The thin plate spline (TPS) or Laplacian surface editing method was used to interpolate a surface and generate a displacement field, which could be calculated by two sets of landmarks located on the patient tissue and normal tissue.
Nevertheless, the accuracy of the obtained landmarks was significantly reduced since the landmarks were located on the pretraumatic optical RGB photos of the patients face.
The method using several landmarks ca not generate detailed point-to-point deformation fields, resulting in a weak ability to generate facial details.

Recently, L. Ma~\cite{malei} proposed a bidirectional face-bone modeling network (P2P-Conv) for orthognathic surgical planning, which uses point-based deep learning methods to encode the bidirectional transformations between facial and bony shapes.
Specifically, the bidirectional point-to-point network (P2P-Net)~\cite{p2p} and the dynamic pointwise convolution (PointConv)~\cite{WuWenxuan} were combined to extract local-to-global spatial information and obtain a bidirectional transformation from the downsampled point cloud.
This method yields substantially improved prediction accuracy.
J. Bao~\cite{JBao} proposed 
P2P-ConvGC
, which is a novel bidirectional deep learning network that includes a module named ConvGC to enhance its contextual learning ability.
Overall, the previously developed networks~\cite{p2p,malei,JBao} rely on efficient point set sampling and the KNN method to extract local features, resulting in a limited receptive field.
Additionally, these networks can handle no more than 5000 points due to GPU memory limitations, for example 4096 points in~\cite{malei}; thus, they are not suitable for addressing dense point clouds.
To overcome these limitations, we propose a Transformer-based network for extracting global information and handling dense points.

\subsection{Point-Based 3D Understanding}
Point-based methods are mainstream deep learning architectures for understanding 3D point clouds that can directly process unstructured points without additional data preprocessing steps~\cite{OACNNs}.
PointNet~\cite{QiCharlesR}, which was developed in a seminal work, combines multilayer perceptrons (MLPs) with a maximum pooling layer to form feature extractors for point coordinates.
PointNet++~\cite{QiCharlesRplus}, which was presented in a follow-up work, was proposed to learn local-global features with hierarchical multiscale perception.
To strengthen the ability to aggregate local neighborhood information, Y. Wang~\cite{WangYue} constructed a graph from neighborhood points and used graph-based convolutions (EdgeConv). 
Several works~\cite{ZhaoHengshuangweb, LanShiyi} have continued to strengthen the degree of local neighborhood utilization.
Recently, Transformer-based architectures~\cite{ZhaoHengshuang,WuXiaoyangv3,YuXumin} have been quickly adapted for 3D understanding.
The pioneering approaches~\cite{GuoMengHaoPCT, ZhaoHengshuang}, similar to the vision Transformer (ViT), directly applied global attention on the given point cloud and local attention between each point and its adjacent points.
These methods are limited by GPU usage and computational complexity constraints due to their unstructured nature.
Thus, additional methods~\cite{WuXiaoyangv2, yangyuswin, liu2023flatformer, WuXiaoyangv3} have been proposed with many designs (including feature embedding, attention and pooling modules) to increase their effectiveness and efficiency.
Recent works~\cite{liu2023flatformer, WuXiaoyangv3} have transformed unstructured, irregular point clouds into standardized sequences according to their spatial proximity and sorted them; these approaches are categorized as serialization-based methods.
These methods, especially PTv3~\cite{WuXiaoyangv3}, demonstrate simpler, faster and stronger performance in indoor and outdoor segmentation and object detection scenarios.
Inspired by them, we propose TCFNet, which combines global attention and local neighborhood information to balance the efficiency achieved when processing dense point clouds and the local deformation of critical organs.
\subsection{Distortion Quantification for Point Clouds}
\begin{figure*}[!t]
	\centerline{\includegraphics[width=\textwidth]{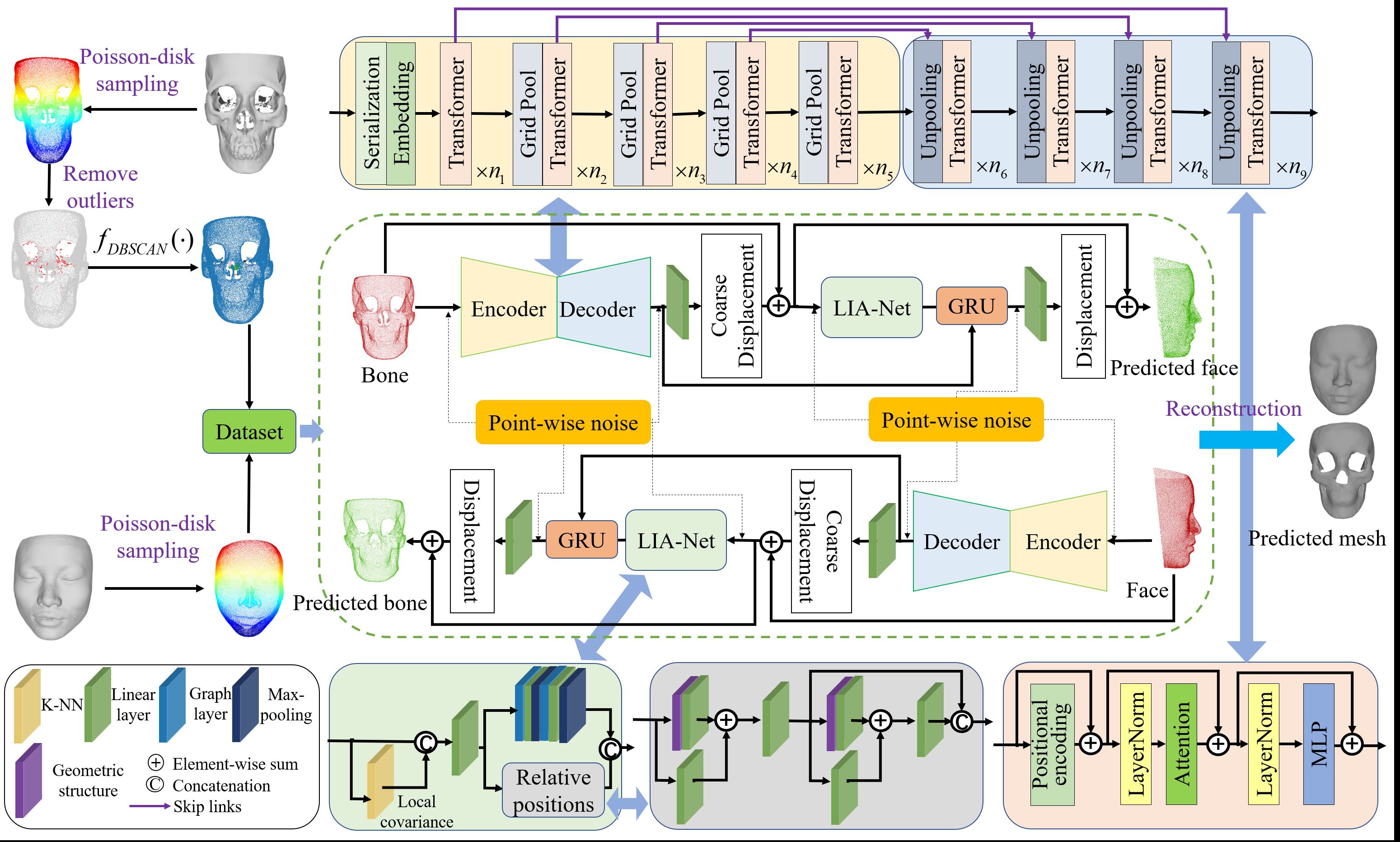}}
	\caption{The framework of our proposed TCFNet.
		It includes a data preprocessing module, a point Transformer network in the first stage and an LIA-Net model in the second stage.
		LIA-Net is composed of local region features, orientations and globally relative position features.
		The pointwise noise is an independent Gaussian noise vector, which is similar to that used in P2P-Net~\cite{p2p}.
		The GRU is a gated recurrent unit.
		The positional encoding scheme is an enhanced conditional positional encoding (xCPE) method~\cite{wang2023octformer,WuXiaoyangv3}.}
	\label{fig1}
\end{figure*}
The Chamfer distance (CD)~\cite{p2p} and Earth mover's distance (EMD)~\cite{achlioptas2018learning} are two acknowledged quality quantification metrics that utilize pointwise distances to measure the similarity between point clouds.
They can constrain shape differences in various point cloud tasks, such as shape reconstruction~\cite{p2p} and completion~\cite{PMPNet, YuXumin}.
Distortion quantization methods must satisfy various requirements including distortion discrimination, differentiability and low computational complexity.
Although the EMD is considered more suitable for maximizing visual quality than the CD is~\cite{WuTong2021}, its expensive computational process and worse descriptions of local point distributions limit its applicability.
The CD is more faithful to local point distributions and is applied in reconstruction, completion and transformation tasks. 
The CD sometimes suffers from being insensitive to different density distributions and outliers.
To address this problem, several works~\cite{WuTong2021, p2p} have proposed density-aware CD methods for measuring the disparities between density distributions.
Furthermore, some researchers~\cite{huang2022learning, huang2023learning} have used learning-based methods to learn to measure shape differences, but these methods have difficulty addressing dense point clouds because of their high GPU memory consumption levels and intensive computations.
Recently, Q. Yang~\cite{YangQiMPED} proposed the multiscale potential energy discrepancy (MPED) measure that uses point potential energy to describe point cloud features.
In summary, we combine the EMD and density distribution of points to compensate for the shortcomings of CD when training our method in this paper.
And these distortion quantification metrics are employed to evaluate algorithm performance.

\section{Method}
The task of conducting bidirectional prediction between faces and bones (i.e., between a facial point cloud $\mathbf{P}_F$ and a corresponding skeletal point cloud $\mathbf{P}_B$) is formulated as a point cloud deformation process.
This process aims to obtain the associated point-to-point displacement fields $\phi_{F\leftrightarrow B}$, which can transform into each other.
The task can be described as
\begin{equation}\begin{split}&\widehat{\phi}_{F\leftrightarrow B}=\mathop{\arg\min}\limits_{\phi_{F\leftrightarrow B}}\mathcal{L}_{sim}\big( \mathbf{P}_F, \mathbf{P}_B \circ \phi_{B\rightarrow F}\big) +\\&\mathcal{L}_{sim}\big(\mathbf{P}_B, \mathbf{P}_F \circ \phi_{F\rightarrow B}\big)+\lambda\mathcal{L}_{reg}\big(\mathbf{P}_F, \mathbf{P}_B, \phi_{F\leftrightarrow B}\big) ,\label{eq1}\end{split}\end{equation}
where $\mathcal{L}_{sim}$ is the distance between the predicted and ground-truth point clouds, which can be used to measure their similarity.
$\mathcal{L}_{reg}$ can overcome the unbalanced optimization of two displacement fields and enhance the transformations to obtain a more uniform distribution~\cite{p2p}.
$\lambda$ is a hyperparameter that can weight the regularization loss.
$\circ$ is the displacement field that warps the input point cloud.

In this paper, we propose a TCFNet model to learn the unique correspondences at the patch level and point level for completing dense point cloud transformations, and the TCFNet structure is shown in Fig. \ref{fig1}.
The overall framework still follows the bidirectional and cyclical structure of P2P-Net~\cite{p2p}, which is an unsupervised shape completion/generation framework without labeling.
The displacement fields generated by one-step deformation methods are more challenging to handle because of their unknown geometries without any other prior information.
Inspired by the existing multistep frameworks~\cite{PMPNetplus, PMPNet, YuXumin}, we adopt a two-step strategy that operates from coarse to fine to realize gradual deformation from in a whole-to-part manner.
Specifically, the first step of the network is based on PTv3~\cite{WuXiaoyangv3}, which can extract the global features of two point clouds and efficiently obtain coarse point movement paths $\phi_{B \rightarrow \widehat{F}}, \phi_{F\rightarrow \widehat{B}}$, thus generating a coarse deformation result $\mathbf{\widehat{H}}_F,\mathbf{\widehat{H}}_B$.
To highlight the local shapes of the target face and bone, LIA-Net is proposed to enhance the local feature representations in the second step, which can promote the local correspondence between the coarse point cloud and the fine point cloud.
It is formulated as follows:
\begin{equation}\mathbf{\widehat{H}}_F=\mathbf{P}_B \circ \phi_{B \rightarrow \widehat{F}}, \mathbf{H}_F=\mathbf{\widehat{H}}_F \circ \phi_{\widehat{F}\rightarrow F}\label{eq2}\end{equation}
\begin{equation}\mathbf{\widehat{H}}_B=\mathbf{P}_F \circ \phi_{F\rightarrow \widehat{B}}, \mathbf{H}_B= \mathbf{\widehat{H}}_B \circ \phi_{\widehat{B}\rightarrow B}\label{eq3}\end{equation}
where $\phi_{\widehat{F}\rightarrow F}, \phi_{\widehat{B}\rightarrow B}$ represents the fine displacement field from the coarse point cloud to the fine point cloud.
$\mathbf{H}_F, \mathbf{H}_B$ are the predicted facial and skeletal point clouds, respectively.

The method can be used in orthognathic surgery planning (from the corrected bone tissue to predicted facial appearance) to predict post-operative recovery status. 
And this method can be employed in reversed orthognathic surgery planning (from the reconstruction facial mesh to the normal bone tissue) to assist the doctor in determining the optimal positions of osteotomized bones. 
The reconstruction facial mesh can be generated from the peritraumatic optical RGB photos of the patient's face.
In the remainder of this section, first, the point Transformer network is described.
Our proposed LIA-Net is then described in detail.
Finally, the utilized loss functions are introduced.

\subsection{Point Transformer Network}
In the first step, a point Transformer network is employed to extract the dependence of each patch feature contained in the input point clouds.
The network combines PTv3~\cite{WuXiaoyangv3} and P2P-Net~\cite{p2p} to generate coarse displacements, and the combined structure consists of noise augmentation, point cloud serialization, pooling, scalar attention and positional encoding operations.
Noise augmentation allows the model to increase the degrees of freedom for the displacement field and learn rich transformations.
Specifically, the pointwise features of the input point or the features of this network output are concatenated with the fixed channels and the same number of pointwise Gaussian noise vectors.
To obtain structured data, point cloud serialization, which is based on space-filling curves, passes through every point; then, the conversion $\varphi$ from one- to $n$-dimensional space is realized.
Furthermore, an inverse mapping $\varphi^{-1}$ is obtained and can be formalized as follows:
\begin{equation}\varphi:\mathbb{Z}\rightarrow \mathbb{Z}^{n}, \varphi^{-1}:\mathbb{Z}\leftarrow \mathbb{Z}^{n}\label{eq4}\end{equation}

Z-order ~\cite{wang2023octformer} and Hilbert curves~\cite{chen2022efficient} are typical space-filling curves that can pass through every point.
Since the neighboring points in the obtained structured data are also spatially close, the reordered variants (with y- or x-axis priority) of the standard space-filling curves can capture diverse local point cloud features.
Thus, the locality-preserving quality of these curves can be leveraged to achieve dimensional compression $\varphi^{-1}$ from a point coordinate $p \in \mathbb{R}^{3}$ into a serialization code $p_i \in \mathbb{N^+}$ with a 64-bit integer using a grid size of $g$.
Finally, the sorting index $p_i$ of the point clouds ordered in this serialization output can obtain local neighbor points, similar to KNN.
The proposed approach is simpler and more efficient than KNN.

Given structured data, the efficient window (patch) and scalar attention mechanisms~\cite{WuXiaoyangv3,liu2023flatformer,yangyuswin} are adopted to calculate the underlying spatial global relationships.
The point cloud serialized with $p_i$ is grouped by the window size $S$ into patches, and it is written as follows:
\begin{equation}\underbrace{p_1,p_2,p_3,p_4}_{\text{window index}},\underbrace{p_5,p_6,p_7,p_8}_{\text{window index}},\underbrace{p_9,p_{10},p_{11},p_{12}}_{\text{window index}},...\label{eq5}\end{equation}
where $S=4$ in Eq. \ref{eq5}.
When the window size exceeds the number of remaining points, the adjacent points in the previous window are used to pad the last window.
Unlike the window interaction scheme of Swin-T~\cite{liu2021swin,zhang2024utsrmorph}, the different sorting order patterns (Z-order, Trans Z-order, Hilbert, and Trans Hilbert patterns) in the point cloud serialization process are shuffled to obtain diverse patterns with rich receptive fields.

To achieve scalar attention, a linear MLP projection is applied to map the features into three vectors (queries $\bm{q}_i$, keys $\bm{k}_i$, and values $\bm{v}_i$) with $c_a$ channels.
Then, the scalar attention $f_{i}^{\text{attn}}$ imposed on one patch $\bm{x}_i$ and its reference patch $\bm{x}_j$ is formulated as
\begin{equation}\bm{w}_{ij}=\langle \bm{q}_i,\bm{k}_j \rangle /\sqrt{c_a},\label{eq6}\end{equation}
\begin{equation}\bm{f}_{i}^{\text{attn}}=\sum\limits_{\bm{x}_j}\text{Softmax}(\bm{w}_{i})_j\bm{v}_j\label{eq7}\end{equation}

The traditional sampling-based pooling method depends on farthest-point or grid sampling.
The sampled points are used to search for and aggregate neighboring point information, but this approach ignores the density of the sampled points and leads to overlaps among different neighbor partitions.
To address this problem, the window-based pooling method is proposed to downsample a subset point $\mathcal{M}_i=(\mathcal{P}_i,\mathcal{F}_i)$ to a point $(\bm{p}^{\prime}_i,\bm{f}^{\prime}_i)$ as follows:
\begin{equation}\bm{f}^{\prime}_i=\text{MaxPool}(\{\bm{f}_j\bm{U}|\bm{f}_j\in \mathcal{F}_i\})\label{eq8}\end{equation}
\begin{equation}\bm{p}^{\prime}_i=\text{MeanPool}(\{\bm{p}_j|\bm{p}_j\in \mathcal{P}_i\})\label{eq9}\end{equation}
where $\mathcal{P}_i,\mathcal{F}_i$ are the position subset and feature subset, respectively.
$(\bm{p}^{\prime}_i,\bm{f}^{\prime}_i)$ are the coordinates of the point and the features.
$\bm{U}$ is a linear projection layer.
Unlike the interpolated upsampling operation in unpooling, the point coordinates are recorded during the pooling process, and the mean point feature is mapped to all point features contained in the same subset.
These modules are flexibly stacked to expand their receptive field and efficiently extract moving features.
Then, the moving features $\bm{f}_{\text{global}}$ and pointwise noise are concatenated and mapped as a point-to-point displacement field, i.e., a coarse deformation displacement field $\phi_{B \rightarrow \widehat{F}}, \phi_{F\rightarrow \widehat{B}}$, which is employed to warp the face-bone point cloud $\mathbf{P}_B, \mathbf{P}_F$ and generate preliminary prediction results $\mathbf{\widehat{H}}_F, \mathbf{\widehat{H}}_B$.

\subsection{Local Information Aggregation Network}
\begin{figure}[!t]
	\centerline{\includegraphics[scale=0.4]{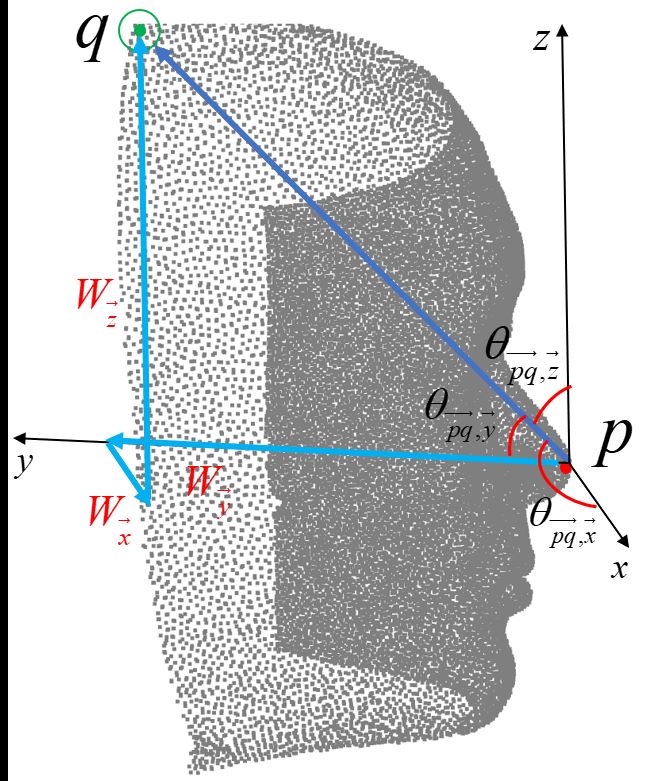}}
	\caption{Capturing the orientation and relative position features between a point $\bm{p}$ and the farthest region $\bm{q}\in \Omega_\text{far}$ via LIA-Net.
		$\theta$ is the orientation between vector $\vec{\bm{pq}}$ and the $x,y,z$-axis. $\bm{W_A}=\bm{W_{\vec{x}}}, \bm{W_{\vec{y}}}, \bm{W_{\vec{z}}}$ denotes learnable weights that can extract direction features from the edge features.}
	\label{fig2}
\end{figure}
The search accuracy of PTv3 in terms of extracting neighbor points is slightly lower than that of KNN, but its local face-bone structures are significant, especially for crucial organs.
Thus, LIA-Net is proposed to promote local tissue correspondences in the second step.
The existing networks~\cite{WangYue, ZhaoHengshuangweb, LanShiyi} extract local features by focusing on modeling local geometric structures.
Some a priori knowledge obtained from the face and bone point clouds is that the relative positions between the local and global regions are fixed.
Inspire by them, our proposed network combines local region features, their orientations, and the globally relative position features.

First, the local covariance matrix for the coarse deformation result is computed via KNN ($K=8$); this matrix is concatenated with the point features and pointwise noise.
Then, we use two branches to model the local structures and their global relative positions.
One branch modifies EdgeConv, which is based on dynamic graphs, into a static edge convolution to maintain consistency within localized regions.
The static edge convolution for a point $(\bm{p},\bm{f})$ is formulated as
\begin{equation}\bm{f}^{\prime}=\text{MaxPool}(\text{MLP}(\{\bm{f_q}-\bm{f}|\bm{q}\in \Omega\}))\label{eq10}\end{equation}
\begin{equation}\bm{f}^{\prime}_{\text{local}}=\text{MaxPool}(\text{MLP}(\{\bm{f^{\prime}_q}-\bm{f}^{\prime}|\bm{q}\in \Omega\}))\label{eq11}\end{equation}
where $\Omega$ is the neighboring region of $\bm{p}$ and $\bm{q}$ is a neighboring point.
This branch can incorporate local neighborhood information.
The other branch can capture the globally relative positions between a point $(\bm{p},\bm{f})$ and the farthest region $\Omega_\text{far}$, as shown in Fig. \ref{fig2}.
To model these relative positions, the points $\{ (\bm{q},\bm{f}_{\bm{q}}),|\bm{q}\in \Omega_\text{far}\}$ that are farthest away from point $(\bm{p},\bm{f})$ are searched in the point cloud.
The edge vectors $\mathop{\bm{p}\bm{q}}\limits ^{\rightarrow}$ are projected to vectors in the $x,y,z$-axes independently, preserving their geometric structures.
The relative lengths are represented by the weight matrix $\bm{W}_d$.
\begin{equation}\bm{W}_d=\frac{(r-|\vec{\bm{pq}}|)^2}{\sum\limits_{\bm{q} \in \Omega_\text{far}}(r-|\vec{\bm{pq}}|)^2},\vec{\bm{pq}}=\bm{p}-\bm{q}\label{eq12}\end{equation}
where $r=\theta_r \cdot \text{Max}(|\vec{\bm{pq}}|)$ and $r$ is the farthest distance between $\bm{p}$ and $\bm{q}$. $\theta_r$ is a weighted parameter that is set to 1.1 in this paper.
The corresponding coefficients $\cos^{2}(\theta_{\vec{\bm{pq}}}, \vec{x}),\cos^{2}(\theta_{\vec{\bm{pq}}}, \vec{y}),\cos^{2}(\theta_{\vec{\bm{pq}}}, \vec{z})$ are used to aggregate each direction of information.
\begin{equation}\cos^{2}(\theta_{\vec{\bm{pq}}}, \bm{A})=\arccos^{2}(\frac{\vec{\bm{pq}}}{|\vec{\bm{pq}}|}), \bm{A}={\vec{x},\vec{y},\vec{z}}\label{eq13}\end{equation}
where $\bm{A}$ is a relative coordinate system that sets $\bm{p}$ as its origin.
The edge features are independently extracted by weight matrices $\bm{W_A}=\bm{W_{\vec{x}}}, \bm{W_{\vec{y}}}, \bm{W_{\vec{z}}}$ in each direction.
The modeled geometric structure then weights the edge features to combine them with the relative position features $\bm{f}^{\prime\prime}_{\text{local}}$.
\begin{equation}\bm{f}^{\prime\prime}= \sum\limits_{\bm{q} \in \Omega_\text{far}}\bm{W}_d \sum\limits_{\bm{A}}(\bm{f}_{\bm{q}}-\bm{f})\cdot \bm{W_A} \cdot \cos^{2}(\theta_{\vec{\bm{pq}}}, \bm{A})\label{eq14}\end{equation}
The residual structure is expressed as
\begin{equation}\bm{f}^{\prime\prime}_{\text{local}}= \text{MLP}(\bm{f}^{\prime\prime}) + \text{MLP}(\bm{f})\label{eq15}\end{equation}
This branch is repeated to enhance the features and obtain $\bm{f}^{\prime\prime\prime}_{\text{local}}$,
and the local information is aggregated as
\begin{equation}\bm{f}_{\text{local}}= \text{MLP}[\bm{f}^{\prime}_{\text{local}},\bm{f}^{\prime\prime}_{\text{local}},\bm{f}^{\prime\prime\prime}_{\text{local}}]\label{eq16}\end{equation}

Since the displacement field in the first step is crucial for determining the movement direction and distance of the path formed in the second step, the previous global features can guide the local displacement field shift.
To achieve this goal, a GRU~\cite{PMPNetplus} is employed to infer the next moving features of each point:
\begin{equation}\bm{f}^{1}_{\text{local}}=\text{GRU}(\bm{f}_{\text{local}},\bm{f}_{\text{global}}) \label{eq17}\end{equation}
The local features $\bm{f}^{1}_{\text{local}}$ and pointwise noise are concatenated, and an MLP layer is used to generate a fine deformation displacement field $\phi_{\widehat{F}\rightarrow F}, \phi_{\widehat{B} \rightarrow B}$.
The displacement fields can warp the coarse transformed results $\mathbf{\widehat{H}}_F, \mathbf{\widehat{H}}_B$ to obtain the final predicted point clouds $\mathbf{H}_F, \mathbf{H}_B$.
\subsection{Loss Functions}
The CD, which is a matching-based loss, encourages points from one point set to correspond to the nearest neighbors in another point set.
It can measure the pointwise distance between a source point cloud $\bm{X}$ and a target point cloud $\bm{Y}$, especially through local structural matching.
It is formulated as
\begin{equation}\mathcal{L}_{CD}(\bm{X},\bm{Y})=\frac{1}{\bm{|X|}}\sum\limits_{\bm{x}\in \bm{X}}\min\limits_{\bm{y}\in \bm{Y}}||\bm{x}-\bm{y}||+\frac{1}{\bm{|Y|}}\sum\limits_{\bm{y}\in \bm{Y}}\min\limits_{\bm{x}\in \bm{X}}||\bm{y}-\bm{x}||\label{eq18}\end{equation}
where $\bm{|X|}$ and $\bm{|Y|}$ are the point sizes.
The EMD aims to find a point-to-point bijection and calculate pairwise distances.
It is defined as
\begin{equation}\mathcal{L}_{EMD}(\bm{X},\bm{Y})=\min\limits_{\phi:\bm{X}\rightarrow \bm{Y}}\frac{1}{\bm{|X|}}\sum\limits_{\bm{x \in X}}||\bm{x}-\phi(\bm{x})||\label{eq19}\end{equation}
where $\phi$ is a bijection that can minimize the mean distance between $\bm{X}$ and $\bm{Y}$.
Notably, $\bm{|X|}=\bm{|Y|}$ in the EMD.
Since finding the optimal mapping is computationally expensive, it is not suitable approach for use with dense point clouds.

\subsubsection{Global Distance Loss}
To measure the distortion value of the deformation point cloud, the similarity losses of the two steps are defined as follows:
\begin{equation}\mathcal{L}^{coarse}_{sim}=\mathcal{L}_{CD}(\mathbf{\widehat{H}}_F, \mathbf{P}_F)+\mathcal{L}_{CD}(\mathbf{\widehat{H}}_B, \mathbf{P}_B)\label{eq20}\end{equation}
\begin{equation}\mathcal{L}^{fine}_{sim}=\mathcal{L}_{CD}(\mathbf{H}_F, \mathbf{P}_F)+\mathcal{L}_{CD}(\mathbf{H}_B, \mathbf{P}_B)\label{eq21}\end{equation}
\subsubsection{Regularization Loss}
The cross-regularization~\cite{p2p} loss is used to enhance the two displacement fields and is formulated as
\begin{equation}\begin{split}&\mathcal{L}_{reg}=\mathcal{L}_{CD}([\mathbf{P}_F, \mathbf{\widehat{H}}_B], [\mathbf{P}_B, \mathbf{\widehat{H}}_F])\\&+\mathcal{L}_{CD}([\mathbf{P}_F, \mathbf{H}_B], [\mathbf{P}_B, \mathbf{H}_F])\end{split}\label{eq22}\end{equation}
\subsubsection{Local Distance Loss}
CD calculations cannot respond to local density distribution fluctuations.
To overcome this problem, we consider the point-by-point correspondences and density distribution in a local neighborhood, which are defined as
\begin{equation}\begin{split}&\mathcal{L}_{local}=\sum\limits_{k=1}^{K}\sum\limits_{n=1}^{N}\frac{||\text{d}(\bm{x},\bm{y})-\text{d}(\bm{x},\bm{z})||}{N\times K}\\&+\beta\mathcal{L}_{EMD}(\Omega_{\mathbf{P}}, \Omega_{\mathbf{H}})\quad\bm{x}\in \mathbf{P},\bm{y}\in \Omega_{\mathbf{P}},\bm{z}\in \Omega_{\mathbf{H}}\end{split}\label{eq23}\end{equation}
where $\mathbf{P}$ denotes the face and bone point clouds $\mathbf{P}_B, \mathbf{P}_F$.
$\Omega_{\mathbf{P}}$ and $\Omega_{\mathbf{H}}$ are the neighborhood regions of point $\bm{x}\in \mathbf{P}$ in the input point cloud $\mathbf{P}$ and the deformed point cloud $\mathbf{H}$, respectively.
$N$ and $K$ are the numbers of point clouds and neighboring points, respectively.
$\beta$ is a weighted parameter.
\subsubsection{Optional Auxiliary Loss}
In deformable medical image registration tasks~\cite{zhang2024utsrmorph}, annotated information derived from critical structures/organs, such as segmentation masks, can yield increased registration accuracy.
Inspired by this, an auxiliary loss is proposed to facilitate local critical feature correspondences, which in turn improve the global correspondences.
\begin{equation}\begin{split}&\mathcal{L}_{aux}=\frac{1}{V}\sum\limits_{v=1}^{V}\bigg( \mathcal{L}_{CD}(\widehat{\Omega}^v_{\mathbf{P}_{B}}, \widehat{\Omega}^v_{\mathbf{P}_{F}}\circ \phi_{F\rightarrow B})\\& + \mathcal{L}_{CD}(\widehat{\Omega}^v_{\mathbf{P}_{F}}, \widehat{\Omega}^v_{\mathbf{P}_{B}}\circ \phi_{B\rightarrow F})\bigg) \end{split}\label{eq24}\end{equation}
where $\widehat{\Omega}_{\mathbf{P}_{B}},\widehat{\Omega}_{\mathbf{P}_{F}}$ are the $v$-$th$ labeled local structures in the face-bone point clouds, respectively.
$V$ is the number of labeled organs.
The labeled local structures can be segmentation masks or bounding boxes of organs.
Finally, the overall loss is formulated as
\begin{equation}\mathcal{L}=\mathcal{L}^{coarse}_{sim} +\mathcal{L}^{fine}_{sim} + \lambda\mathcal{L}_{reg}+ \mathcal{L}_{local} + \{\mathcal{L}_{aux}\}_{\text{optional}}\label{eq25}\end{equation}
\section{Experiments}
\subsection{Datasets}
\subsubsection{Data Preprocessing}
\begin{table*}
	\caption{The results produced by our approach and other comparison methods on the test set according to a 5-fold cross-validation process.}
	\setlength{\tabcolsep}{3pt}
	\begin{tabular}{p{50pt}p{30pt}p{25pt}p{50pt}p{50pt}p{50pt}p{50pt}p{60pt}p{40pt}p{30pt}}
		\hline
		Method & Shape&$N_2$& MPED &CD (mm)& EMD (mm)& JSD& HD (mm)& Time (s)& Memory\\
		\hline
		\multirow{3}{*}{P2P-Net}& All& \multirow{3}{*}{4096}& 90.24$\pm$25.19& 4.56$\pm$0.92& 3.62$\pm$0.91& 0.38$\pm$0.1& 35.49$\pm$6.53&\multirow{3}{*}{0.22}&\multirow{3}{*}{15232}\\
		& Bone& & 43.39$\pm$12.23& 2.21$\pm$0.44& 1.93$\pm$0.55& 0.19$\pm$0.06& 15.38$\pm$4.08&&\\
		& Face& & 46.85$\pm$17.67& 2.36$\pm$0.65& 1.69$\pm$0.65& 0.18$\pm$0.07& 20.11$\pm$4.78&&\\
		\hline
		\multirow{3}{*}{P2P-Conv}& All& \multirow{3}{*}{4096}& 82.68$\pm$26.95& 4.27$\pm$0.96& 4.58$\pm$1.58& 0.34$\pm$0.11& 180.8$\pm$172.5&\multirow{3}{*}{1.13}&\multirow{3}{*}{14158}\\
		& Bone& & 37.32$\pm$10.52& 1.97$\pm$0.39& 2.48$\pm$0.58& 0.18$\pm$0.05& 75.76$\pm$108.92&&\\
		& Face& & 45.36$\pm$20.47& 2.3$\pm$0.72& 2.1$\pm$1.21& $\mathbf{0.16\pm0.07}$& 105.04$\pm$140.4&&\\
		\hline
		\multirow{3}{*}{PTv3}& All& \multirow{3}{*}{4096}& 80.64$\pm$26.88& 4.26$\pm$1.01&  $\mathbf{2.94\pm1.27}$& 0.34$\pm$0.12& 35.95$\pm$11.71&\multirow{3}{*}{0.19}&\multirow{3}{*}{$\mathbf{3912}$}\\
		& Bone& & 38.22$\pm$12.27& 2.04$\pm$0.44& $\mathbf{1.47\pm0.62}$& 0.17$\pm$0.06& 16.03$\pm$6.7&&\\
		& Face& & 42.42$\pm$17.72& 2.22$\pm$0.65& $\mathbf{1.48\pm0.71}$& $\mathbf{0.16\pm0.07}$& 19.92$\pm$7.26&&\\
		\hline
		\multirow{3}{*}{PTv3}& All& \multirow{3}{*}{20480}& 82.64$\pm$25.65& 4.3$\pm$0.93& 4.04$\pm$1.56& 0.34$\pm$0.12& 35.25$\pm$9.46&\multirow{3}{*}{$\mathbf{0.05}$}&\multirow{3}{*}{4500}\\
		& Bone& & 38.39$\pm$11.58& 2.03$\pm$0.42& 2.29$\pm$0.93& 0.17$\pm$0.06& 15.27$\pm$4.79&&\\
		& Face& & 44.25$\pm$17.68& 2.26$\pm$0.64& 1.75$\pm$0.78& 0.17$\pm$0.08& 19.98$\pm$6.78&&\\
		\hline
		\multirow{3}{*}{TCFNet}& All& \multirow{3}{*}{20480}& $\mathbf{77.65\pm25.68}$& $\mathbf{4.11\pm0.96}$& 3.6$\pm$1.59& $\mathbf{0.31\pm0.12}$& $\mathbf{33.3\pm9.03}$&\multirow{3}{*}{0.3}&\multirow{3}{*}{22436}\\
		& Bone& & $\mathbf{36.33\pm12.11}$& $\mathbf{1.96\pm0.45}$& 1.88$\pm$0.84& $\mathbf{0.16\pm0.07}$& $\mathbf{14.36\pm4.28}$&&\\
		& Face& & $\mathbf{41.32\pm17.36}$& $\mathbf{2.15\pm0.65}$& 1.73$\pm$0.8& $\mathbf{0.16\pm0.08}$& $\mathbf{18.95\pm6.75}$&&\\
		\hline
		\multicolumn{10}{p{500pt}}{'Shape' is the error of different object.
			'Bone' and 'Face' represent the skeletal and facial errors, respectively.
			'ALL' is the sum of these errors.
			$N_2$ is the number of input points.
			Here, $p<0.05$ indicates that the accuracy of our approach is significantly greater than that of the comparison methods.
		}
	\end{tabular}
	\label{tab1}
\end{table*}
The datasets consist of 120 paired face-bone samples, which include skeletal CT data and corresponding facial CT data derived from healthy volunteers. 
Consistent with the inclusion criteria established by~\cite{malei}, participants were excluded if they had any documented history of CMF trauma, congenital/acquired jaw deformities, or surgical interventions involving the facial/jaw region.
They were retrospectively recruited from the Department of Maxillofacial Surgery that is cooperatively managed by the Peking University School and the Hospital of Stomatology.
The CT volume data was segmented using threshold and manual fine-tuning by a professional surgeon. 
The segmented mask was reconstructed to obtain a surface mesh using the Marching Cube algorithm. 
To evaluate the performance of our method on the local regions of interest (ROIs), such as the nose and lip, these ROIs were labeled with bounding boxes by the surgeon.
The ROIs in the input point clouds were transformed to a local predicted point cloud using the displacement fields generated by the network.
The local errors were then measured as the distance between the local predicted point cloud and the labeled ground-truth.

Dense point clouds ($\mathbf{P}^1_B,\mathbf{P}^1_F$) were sampled from a 3D shape mesh to generate $M_1=20$ point clouds via Poisson disk sampling, one of which contained $N_1=25000$ points.
Since the point clouds contained outliers, the statistical outlier removal method was used to remove outliers, i.e., neighboring points that were further from the average.
For the internal patchy noise points, we adopted the density-based spatial clustering of applications with noise (DBSCAN) method for detection.
This paper adopted the standard point cloud regularization operation, which was implemented as follows:
we subtracted the mean value of each dimension from the coordinates of each data point to exchange the centroid of this point cloud with the origin and then divided the coordinates by the farthest distance between one point and the origin.
Finally, the remaining subsets were sampled by farthest-point sampling (FPS) to $N_2=20480$ points, and then the dense network inputs of $\mathbf{P}_F, \mathbf{P}_B$ were obtained.
The sparse point clouds $N_2=4096$ were obtained by FPS from the dense point cloud.
A total of 120 paired samples were divided into 95 and 25 samples for training and testing, respectively.
$95\times M_1=1900$ dense point clouds were used to train our method, and $95\times 5 \times M_1=9500$ sparse point clouds were used to train several comparison methods due to GPU memory limitations.
\subsubsection{Data Postprocessing}
The predicted dense point clouds were generated by our proposed TCFNet model.
For models requiring an input size of 4,096 points, the original 20,480-point cloud was downsampled four times to generate five distinct 4,096-point subsets. 
Each subset was individually processed through the models, and the resulting five output point clouds were merged to reconstruct the full-resolution 20,480-point cloud. 
All quantitative metrics were subsequently computed using this reconstructed high-density point cloud.
And its normals were estimated by locally fitting a plane for each point.
Then, the normal orientations were oriented toward consistency via a minimum spanning tree.
The smooth surfaces were reconstructed via the Poisson surface reconstruction method~\cite{kazhdan2006poisson}, which views this problem as a regularized optimization problem.
To delete the triangles of the outliers, a density threshold was employed to obtain the final predicted mesh.
Additionally, Taubin smooth was integrated to refine the reconstructed surfaces, ultimately enhancing local detail, overall quality, and smoothness of the predicted mesh. 
Finally, Hausdorff distance metric was utilized to assess the reconstruction error. All post-processing operations were performed in MeshLab.
\subsection{Comparison Methods and Evaluation Metrics}
\subsubsection{Comparison Methods}
The comparison methods are summarized as follows.

(1) \textbf{P2P-Net}~\cite{p2p}, which is a point-based network, can learn bidirectional point-to-point transformations between 3D point clouds from two domains.
Its core is based on PointNet++~\cite{QiCharlesRplus}, which uses many set abstraction and feature propagation layers to capture global features and then generates a displacement field.

(2) \textbf{P2P-Conv}~\cite{malei}, which is a point-based convolutional network, can predict bidirectional point-to-point transformations.
It leverages PointConv to gradually extract features in a local-to-global manner.

(3) \textbf{PTv3}~\cite{WuXiaoyangv3}, which is an efficient point-based Transformer network, can accurately and efficiently process a point cloud in indoor/outdoor semantic segmentation tasks, especially its large receptive field.
In the PTv3 experiments, PointNet++ in P2P-Net was replaced with PTv3.

A geometric loss and cross-regularization were employed to train these single-stage methods~\cite{p2p}.
\subsubsection{Evaluation Metrics}
We introduce several evaluation metrics.

(1) \textbf{CD} is the point-to-point nearest distance.

(2) \textbf{EMD} represents the global correspondence distance.

(3) \textbf{MPED}~\cite{YangQiMPED} uses a multiscale approach to achieve global-local quantification.

(4) \textbf{Jensen-Shannon Divergence (JSD)}~\cite{achlioptas2018learning} can measure the marginal distributions between two point clouds in the Euclidean 3D space via the Kullback-Leibler (KL) divergence.

(5) The \textbf{Hausdorff distance (HD)} defines the maximum mismatch degree.

(6) The \textbf{Wilcoxon rank-sum test} can be employed to assess significant differences between the results of one method and those of other methods.

(7) The \textbf{time cost (s)} is the inference time required for one pair of point clouds.

(8) \textbf{GPU memory usage (MB)}.

Lower CD, EMD, MPED and JSD indices indicated that the two compared point distributions were more proximate.
A larger HD index reflected more pronounced dispersion points in the point cloud after performing deformation.

\subsection{Implementation Details}
The methods implemented in this paper were trained on one workstation with an NVIDIA GeForce RTX 3090Ti GPU and built with Ubuntu 20.04, PyTorch 2.1, and python 3.8.
Our preprocessing and postprocessing methods relied on Open3D 0.18.0 and MeshLab.
The optimizer was Adam, and the batch size was 2.
The initial learning rate was 0.01, and it was reduced to $\frac{1}{10}$ in epochs 30 and 45.
The maximum number of epochs was 60.
The weighted parameters $\beta$ and $\alpha$ in the loss functions were set to 0.1 and 0.3, respectively.
The number of neighboring points $K$ was 16 in the local distance loss.
In the point Transformer network, the patch size was 1024.
The grid size of our data was set to 0.01.
In the encoder, the numbers of heads, Transformer blocks and channels were [2, 4, 8, 16, 32], [2, 2, 2, 6, 2] and [32, 64, 128, 256, 512], respectively.
Similarly, these parameters in the decoder were [4, 4, 8, 16], [2, 2, 2, 2] and [64, 64, 128, 256].
To highlight local structures, the number of neighborhood points used in the KNN operations was 8 for LIA-Net.
\subsection{Results}
\begin{figure*}[!t]
	\centerline{\includegraphics[width=\textwidth]{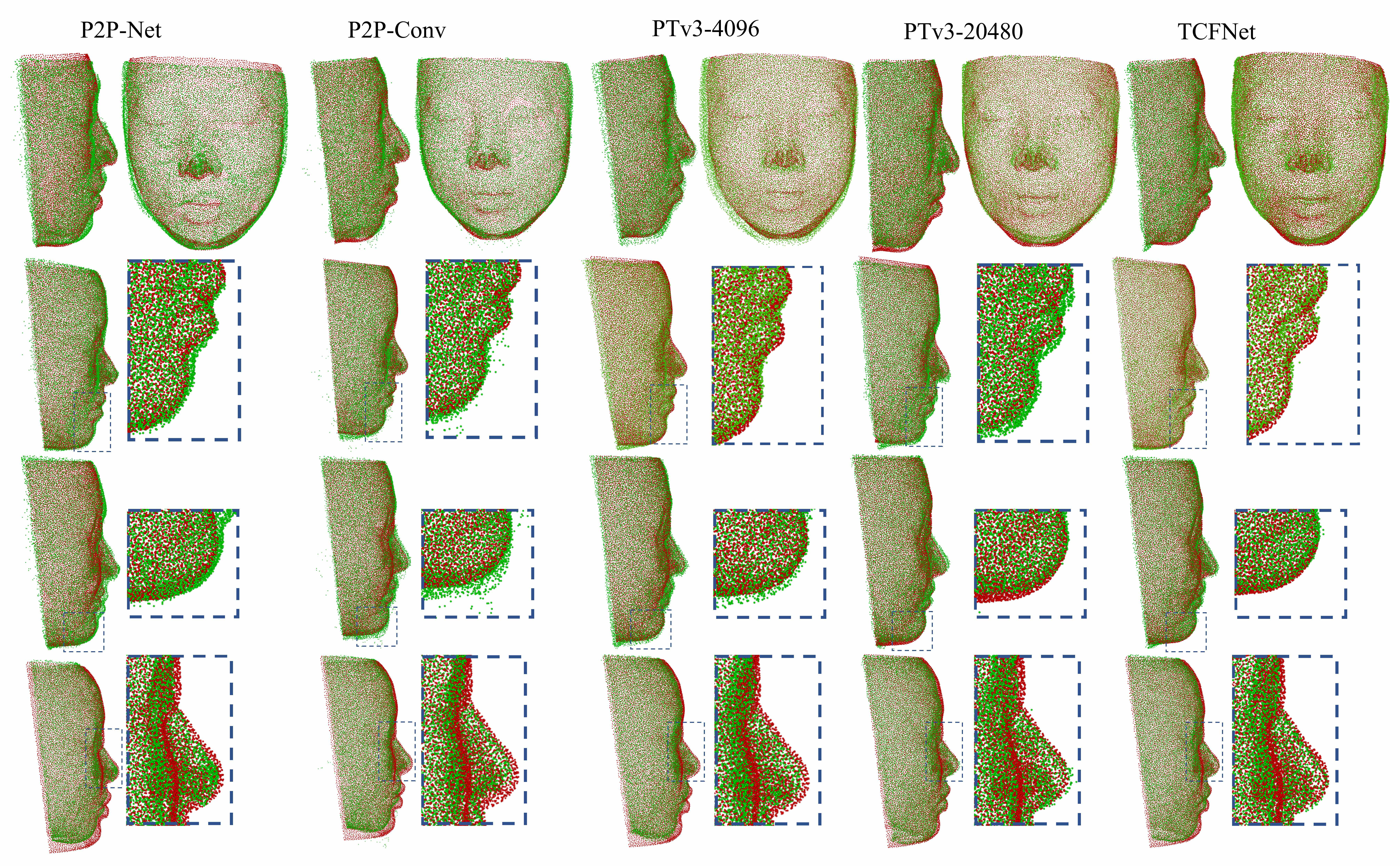}}
	\caption{Comparison between the results produced for the predicted facial point clouds and the ground-truth point clouds. The green points are the predicted points, and the red points are the ground-truth points.}
	\label{fig3}
\end{figure*}
\begin{figure*}[!t]
	\centerline{\includegraphics[width=\textwidth]{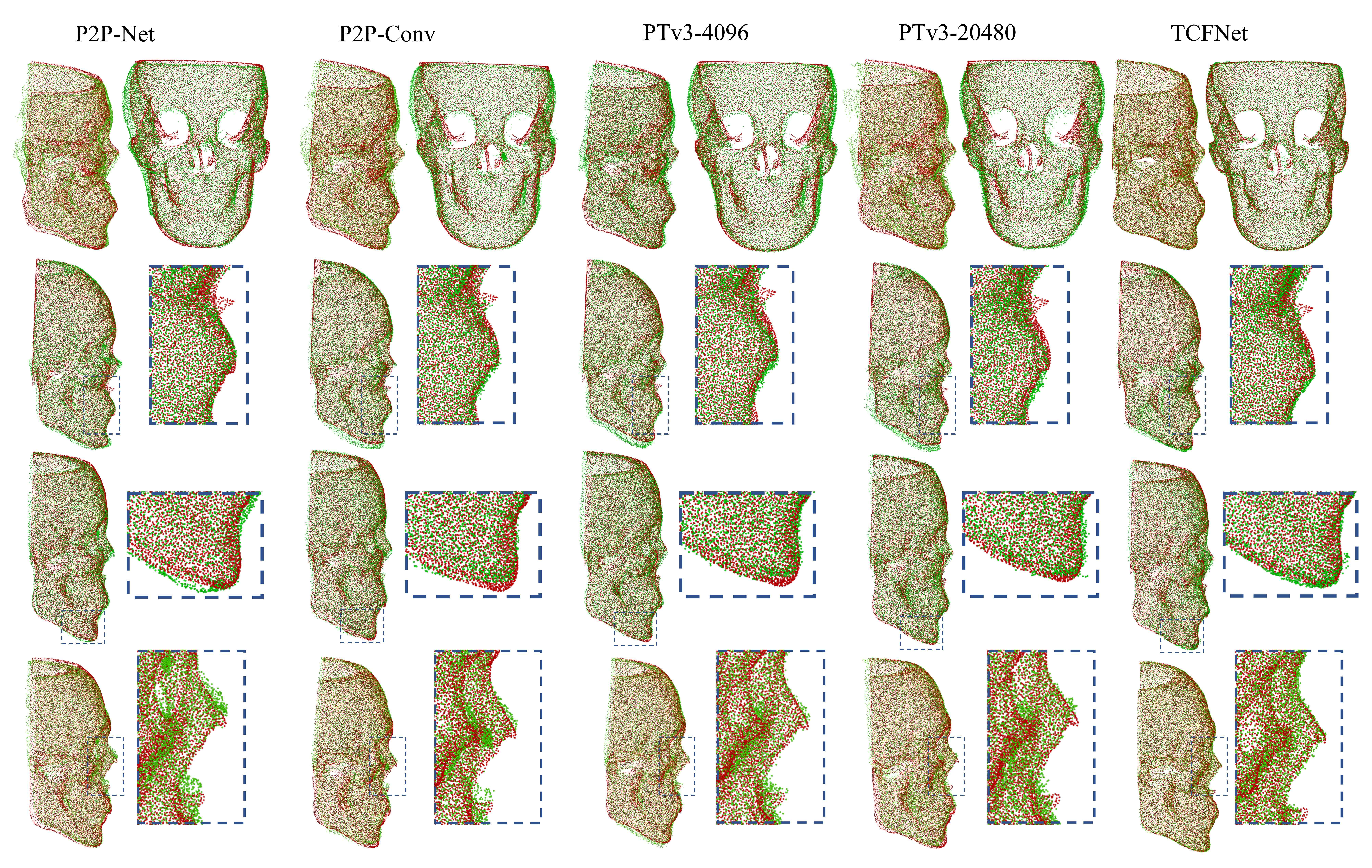}}
	\caption{Comparison between the results produced for the predicted skeletal point clouds and the ground-truth point clouds.}
	\label{fig4}
\end{figure*}
\begin{figure*}[!t]
	\centerline{\includegraphics[width=\textwidth]{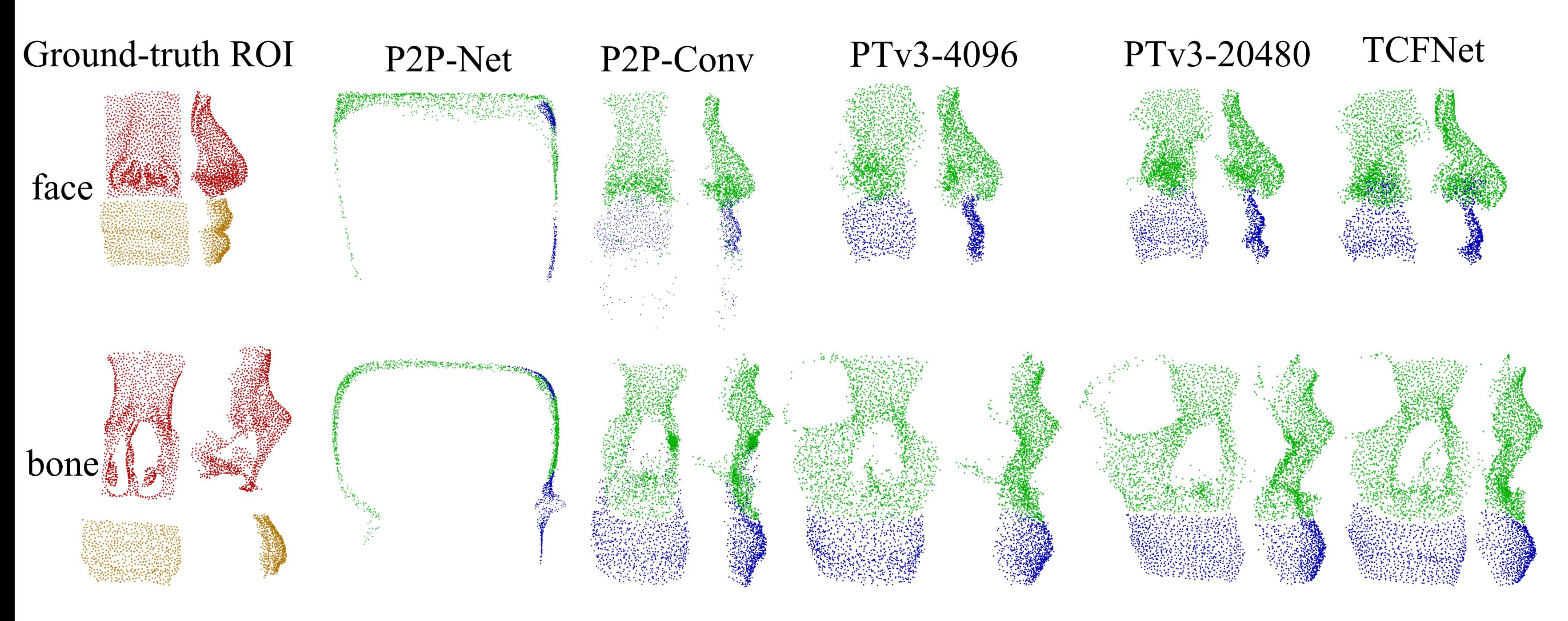}}
	\caption{The visual comparison of the local ROI between results produced for the predicted facial point clouds and the ground-truth point clouds. The predicted local point cloud (face and bone) is obtained from the ground-truth labeled ROI (bone and face) after network transformation, and this bone and face point cloud come from a person. The red, yellow, green, blue points are the ground-truth nose, lip, predicted nose and lip, respectively.}
	\label{fig5}
\end{figure*}
\begin{figure*}[!t]
	\centerline{\includegraphics[width=\textwidth]{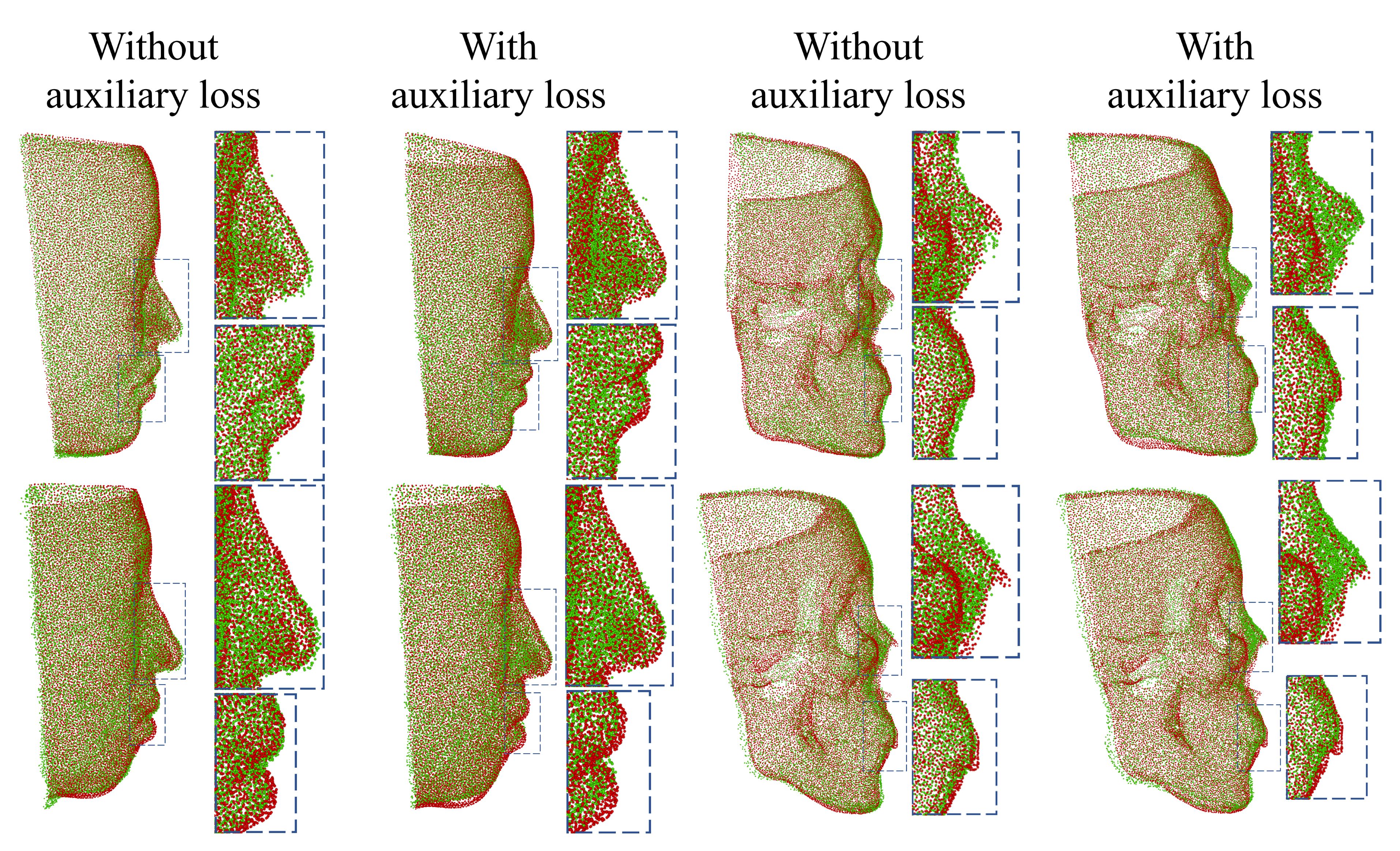}}
	\caption{Comparison between the results produced in local areas for the predicted point clouds and the ground-truth point clouds.}
	\label{fig6}
\end{figure*}
\begin{figure*}[!t]
	\centerline{\includegraphics[width=\textwidth]{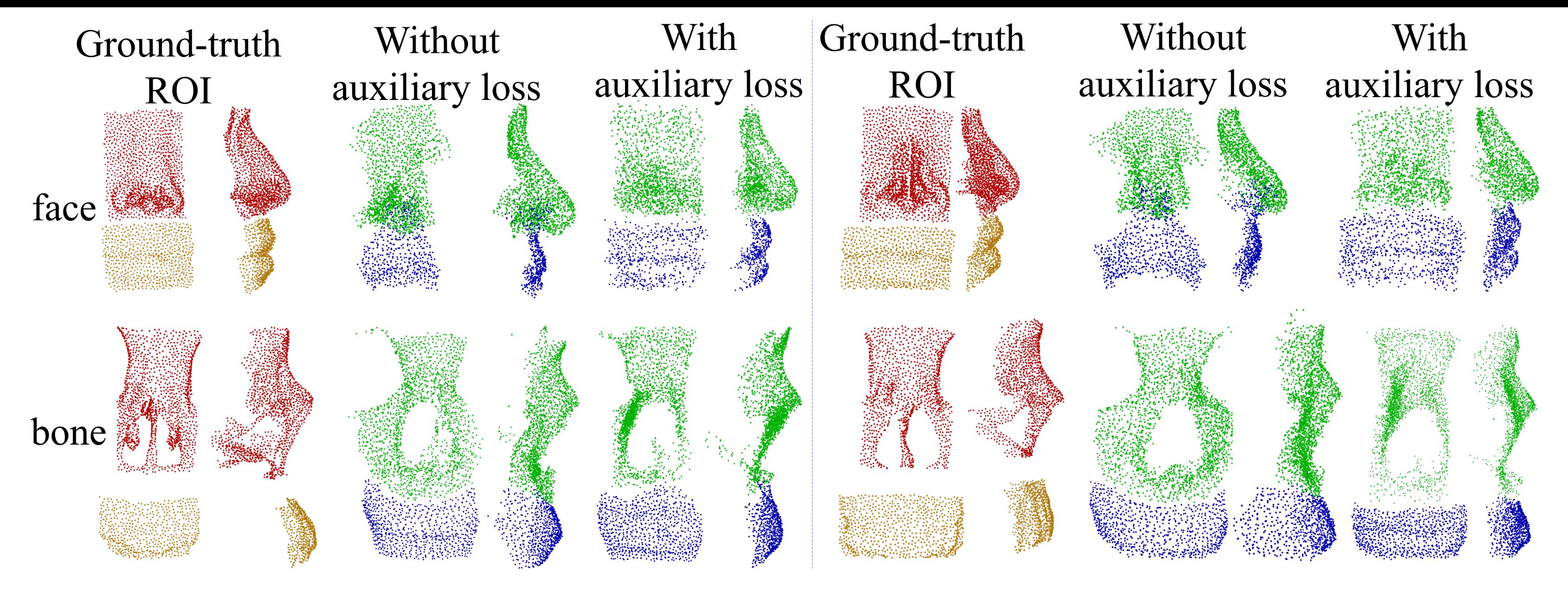}}
	\caption{The visual comparison of the local ROI between results produced for the predicted facial point clouds and the ground-truth point clouds. Individual columns correspond to a pair of bony and facial structures from the same person. The red, yellow, green, blue points are the ground-truth nose, lip, predicted nose and lip, respectively.}
	\label{fig7}
\end{figure*}
\begin{table}
	\caption{The local errors between our approach and the other SOTA methods on the test set, determined via 5-fold cross-validation.}
	\setlength{\tabcolsep}{3pt}
	\begin{tabular}{p{35pt}p{25pt}p{60pt}p{45pt}p{45pt}}
		\hline
		Method & Shape& MPED &CD (mm) & HD (mm)\\
		\hline
		\multirow{3}{*}{P2P-Net}& All& 2519.52$\pm$472.31& 83.13$\pm$15.1& 164.63$\pm$33.22\\
		& Nose& 1212.57$\pm$265.44& 39.92$\pm$8.48& 83.7$\pm$17.77\\
		& Lip& 1306.94$\pm$234.81& 43.21$\pm$7.53& 80.97$\pm$17.58\\
		\hline
		\multirow{3}{*}{P2P-Conv}& All& 146.95$\pm$41.89& 6.13$\pm$1.45& 53.93$\pm$12.83\\
		& Nose& $\mathbf{61.21\pm18.94}$& $\mathbf{2.61\pm0.69}$& 27.15$\pm$9.58\\
		& Lip& 85.74$\pm$29.96& 3.52$\pm$0.97& 26.78$\pm$6.81\\
		\hline
		\multirow{3}{*}{PTv3}& All& 162.04$\pm$50.46& 6.75$\pm$1.68& 46.67$\pm$9.55\\
		& Nose& 91.02$\pm$30.65& 3.7$\pm$1.03& 26.79$\pm$9.49\\
		-4096& Lip&71.02$\pm$25.43& 3.05$\pm$0.87& 19.87$\pm$7.36\\
		\hline
		\multirow{3}{*}{PTv3}& All& 167.81$\pm$51.3& 6.94$\pm$1.67& 51.87$\pm$10.09\\
		& Nose& 91.75$\pm$31.29& 3.71$\pm$1.16& 30.54$\pm$11.6\\
		-20480& Lip& 76.06$\pm$31.21& 3.22$\pm$1.07& 21.33$\pm$8.12\\
		\hline
		\multirow{3}{*}{TCFNet}& All& $\mathbf{139.58\pm 44.29}$& $\mathbf{5.87\pm1.47}$& $\mathbf{43.54\pm10.51}$\\
		& Nose& 70.25$\pm$26.28& 2.93$\pm$0.85& $\mathbf{24.62\pm10.45}$\\
		& Lip& $\mathbf{69.28\pm26.55}$& $\mathbf{2.94\pm0.91}$& $\mathbf{18.92\pm6.63}$\\
		\hline
		\multicolumn{5}{p{240pt}}{'Nose'/'Lip' represent the errors of the predicted bone and face at the nose and lip, respectively.
		Here, $p<0.05$ indicates that the accuracy of our approach is significantly greater than that of the comparison methods.}
	\end{tabular}
	\label{tab2}
\end{table}

\begin{table*}
	\caption{Comparison between the results obtained with the optional auxiliary loss and other losses.}
	\setlength{\tabcolsep}{3pt}
	\begin{tabular}{p{30pt}p{30pt}p{50pt}p{50pt}p{50pt}p{50pt}p{50pt}}
		\hline
		Loss& Shape & MPED &CD (mm) & EMD (mm)& JSD& HD (mm) \\
		\hline
		\multirow{2}{*}{$\mathcal{L}_{local}$}& All& 77.75$\pm$26.8& $\mathbf{3.97\pm1.02}$& 5.81$\pm$1.52& 0.34$\pm$0.12&32.98$\pm$8.71\\
		& Local& 111.9$\pm$34.2& 4.9$\pm$1.18& & &38.73$\pm$13.31\\
		\hline
		\multirow{2}{*}{$\mathcal{L}_{EMD}$}& All& 76.14$\pm$29.9& 4.07$\pm$1.12& 3.57$\pm$1.32& 0.32$\pm$0.13&$\mathbf{31.48\pm7.96}$\\
		& Local& 129.4$\pm$38& 5.62$\pm$1.34& & &44.82$\pm$14.03\\
		\hline
		\multirow{2}{*}{TCFNet}& All& $\mathbf{75.08\pm30.46}$& 4.04$\pm$1.14& $\mathbf{3.39\pm1.39}$& $\mathbf{0.31\pm0.13}$&31.5$\pm$7.95\\
		& Local& 122.49$\pm$36.35& 5.38$\pm$1.25& & &33.87$\pm$8.07\\
		\hline
		\multirow{2}{*}{$\mathcal{L}_{aux}$}& All& 77.86$\pm$29.1& 4.09$\pm$1.09& 4.23$\pm$1.3& 0.33$\pm$0.13&31.99$\pm$7.47\\
		& Local& $\mathbf{85.36\pm27.7}$& $\mathbf{4.08\pm1.03}$& & &$\mathbf{21.95\pm7.55}$\\
		\hline
		\multicolumn{7}{p{350pt}}{'All' is the sum of the bone and face prediction errors.
			'Local' is the sum of the nose and lip errors in the bone and face.
			$\mathcal{L}_{local}$ means to remove the local distance loss, and $\mathcal{L}_{EMD}$ represents removing the EMD module.  
			$\mathcal{L}_{aux}$ represents the auxiliary loss that is added to our method.}
	\end{tabular}
	\label{tab3}
\end{table*}
\begin{table*}
	\caption{The results produced by our proposed method in ablation experiments.}
	\setlength{\tabcolsep}{3pt}
	\begin{tabular}{p{80pt}p{40pt}p{60pt}p{50pt}p{50pt}p{50pt}p{50pt}}
		\hline
		Module& Shape & MPED &CD (mm) & EMD (mm)& JSD& HD (mm) \\
		\hline
		\multirow{2}{*}{LIA-Net}& All& 80.72$\pm$29.39& 4.24$\pm$1.09& 3.74$\pm$1.24& 0.34$\pm$0.13& 31.56$\pm$8.5\\
		& Local& 151.35$\pm$40.68& 6.4$\pm$1.35& & &49.24$\pm$13.38\\
		\hline
		\multirow{2}{*}{GRU}& All& 80.11$\pm$31.55& 4.17$\pm$1.18& 3.85$\pm$1.53& 0.33$\pm$0.14& 32.36$\pm$6.96\\
		& Local& 136.18$\pm$38.6& 5.81$\pm$1.33& & & 37.93$\pm$8.26\\
		\hline
		\multirow{2}{*}{Relative positions}& All& 79.93$\pm$29.08& 4.16$\pm$1.09& 4.53$\pm$1.44& 0.33$\pm$0.12& 33.72$\pm$11.1\\
		& Local& 134.93$\pm$36.65& 5.77$\pm$1.25& & & 43.01$\pm$12.24\\
		\hline
		\multirow{2}{*}{TCFNet}& All& $\mathbf{75.08\pm30.46}$& $\mathbf{4.04\pm1.14}$& $\mathbf{3.39\pm1.39}$& $\mathbf{0.31\pm0.13}$& $\mathbf{31.5\pm7.95}$\\
		& Local& $\mathbf{122.49\pm36.35}$& $\mathbf{5.38\pm1.25}$& & &$\mathbf{33.87\pm8.07}$\\
		\hline
		\multicolumn{7}{p{420pt}}{'Module' denotes the removal of the corresponding module from our proposed method.  }
	\end{tabular}
	\label{tab4}
\end{table*}
\subsubsection{Comparison with the SOTA Methods}
The state-of-the-art (SOTA) methods are compared in Table~\ref{tab1}, showing an entire comparison of the whole face and bone between the predicted transformation and the ground truth.
And the comparison among the local error results is shown in Table~\ref{tab2}, which was calculated by comparing the nose and lip regions between the predicted transformation and the ground truth.
Our method achieved the lowest error, except in terms of the EMD metrics, which were slightly larger than those of PTv3-4096.
This reflects the notion that our method can realize global transformations while considering local structural correspondences and hence achieve excellent performance in face-bone transformation tasks involving dense point clouds.

The whole visualization results are shown in Fig. \ref{fig3}, \ref{fig4}.
The predicted results obtained by our proposed method more closely matched the ground truth and had fewer outliers.
The visualization result of the nose and lip regions is shown in Fig.~\ref{fig5}.
Our proposed method further improves one-to-one organ correspondence while reducing noise, demonstrating outstanding performance compared to SOTA methods.
Among the other SOTA methods, P2P-Net yielded abnormal outputs, leading to significant errors.
This indicates that the displacement field of P2P-Net fails to establish a one-to-one correspondence between local structures. 
We speculate that this may be attributed to PointNet++'s the farthest point sampling strategy and upsampling interpolation method. 
The displacement fields generated by other methods successfully establish correspondences in local structures. 
P2P-Conv resulted in suboptimal MPED and CD metrics, as shown in Table~\ref{tab2}, which indicates that P2P-Conv can yield better performance for local structures because of the local feature extraction ability of pointwise convolution; however, it produces many outliers and decreases global point-to-point correspondences.
Although PTv3 achieved a better EMD metric for global correspondences than did other methods, it significantly reduced the accuracy of the local structure, which is a primary concern for surgeons.
Notably, the accuracy of PTv3 decreased significantly when the number of points increased, probably because a decrease in its ability to search in the neighborhood of each point was exacerbated by discarding the KNN operation.
Thus, our proposed LIA-Net compensated for the inferior deformation of local areas in PTv3 by modeling local geometric structures.

The statistical analysis for the Chamfer distance (CD) indicator is employed to assess significant differences between our proposed method and other baselines, and the CD indicator is calculated in the test set via 5-fold cross-validation from Table \ref{tab1}, \ref{fig2}. 
The $p$ value of our proposed method compared with other methods is $p<0.05$. 
It indicates that the accuracy of our approach is significantly greater than that of the competing methods.
\begin{figure*}[!t]
	\centerline{\includegraphics[width=\textwidth]{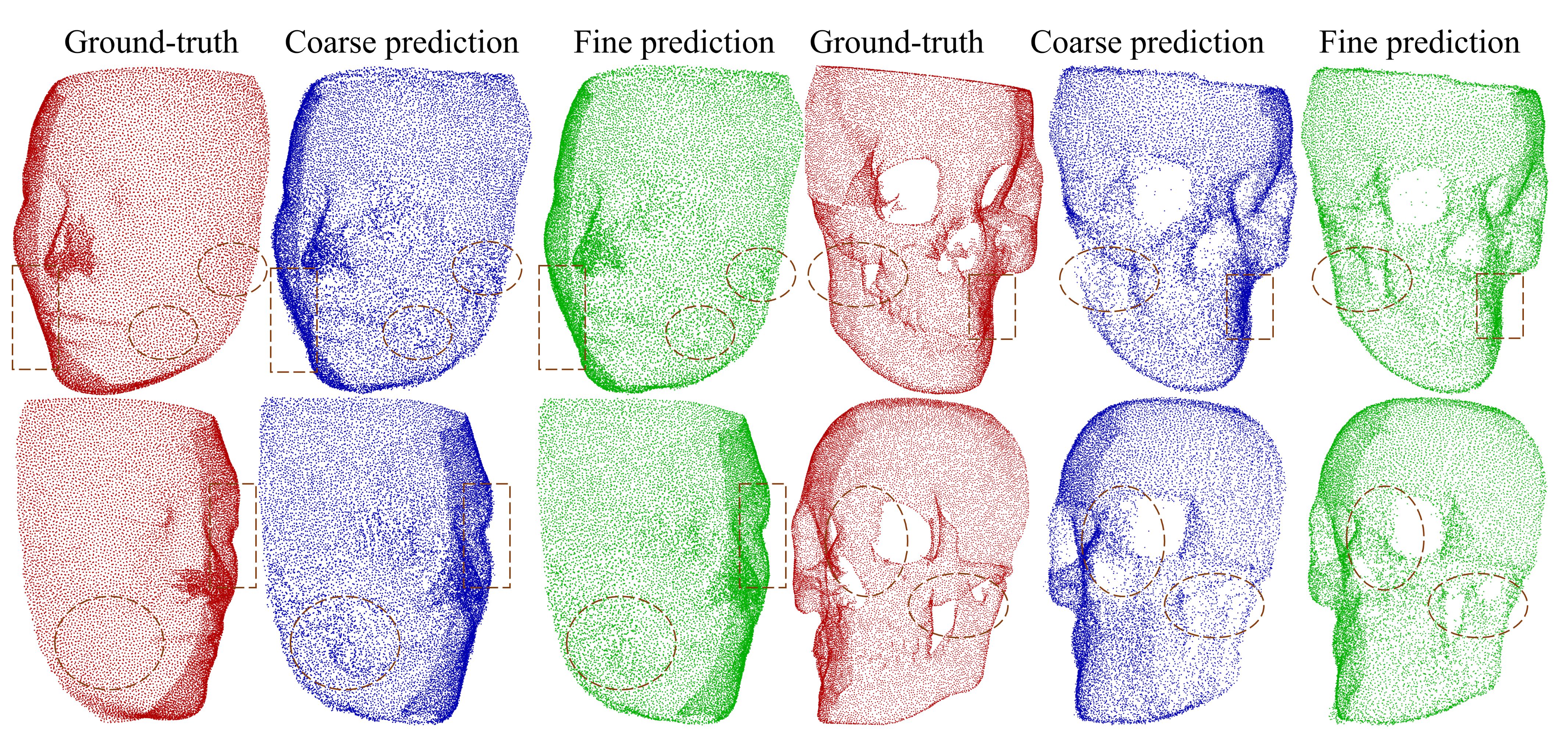}}
	\caption{Visualization results of coarse and fine predicted point clouds of our method.}
	\label{fig8}
\end{figure*}
\begin{figure*}[!t]
	\centerline{\includegraphics[width=\textwidth]{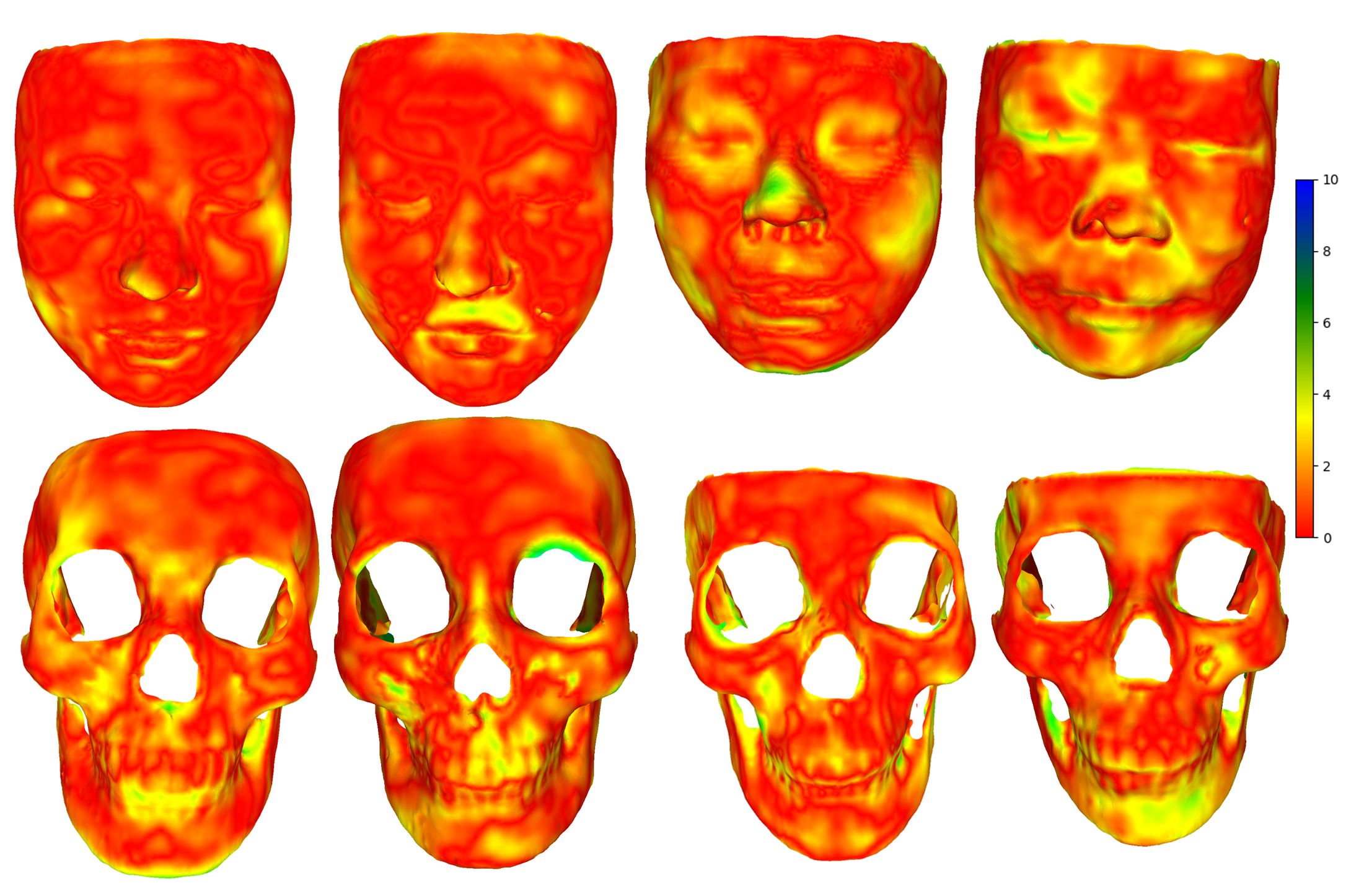}}
	\caption{Comparisons of the error distribution between the predicted and ground-truth meshes from 4 persons in test set.}
	\label{fig9}
\end{figure*}
\subsubsection{Effectiveness of the Auxiliary Loss}
\begin{figure*}[!t]
	\centerline{\includegraphics[scale=0.3]{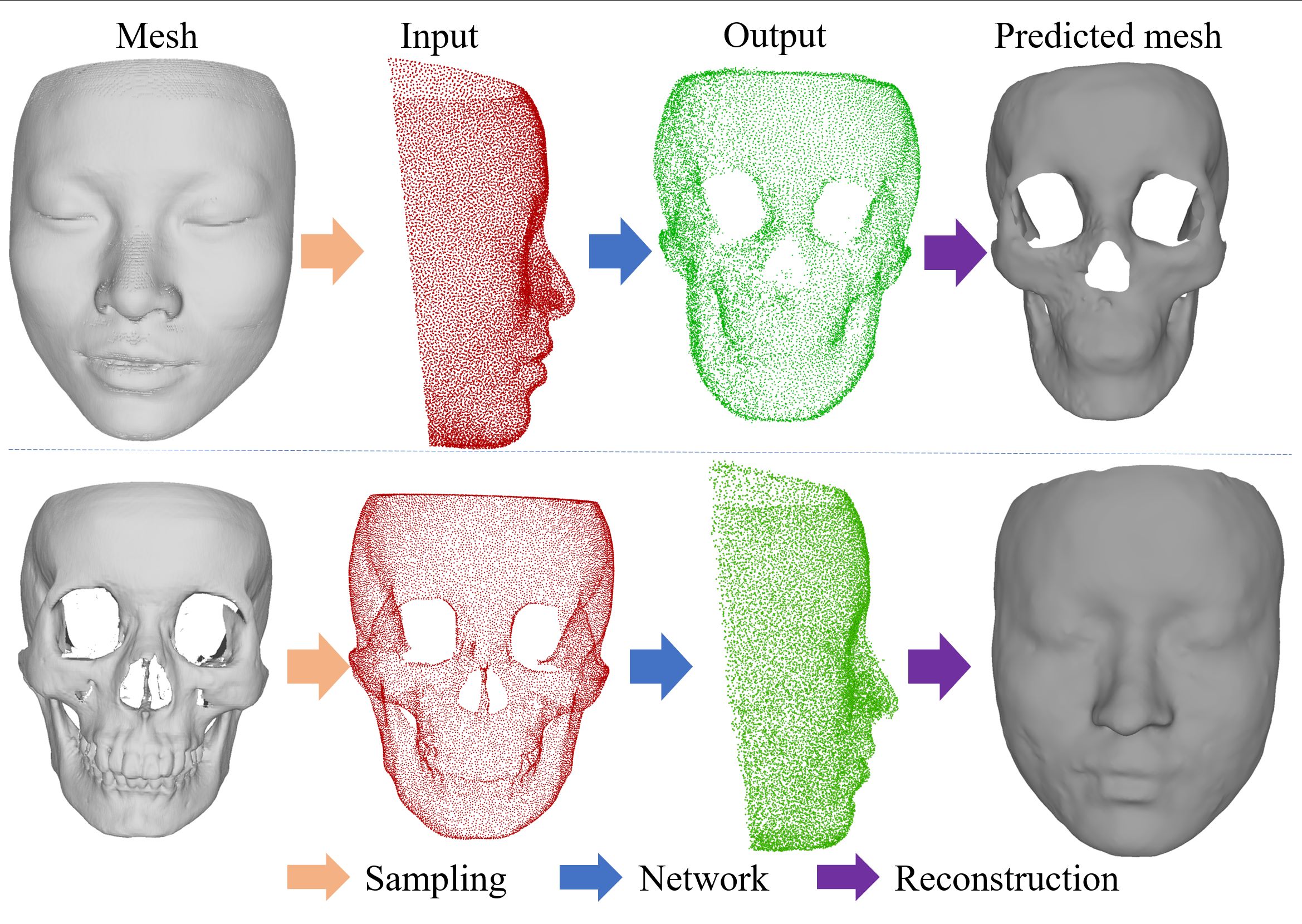}}
	\caption{The prediction results of our method. The mesh is sampled into a point cloud which is the input of our network. The output of the point cloud is reconstructed to a mesh for better visualization. The data is from the same person.}
	\label{fig10}
\end{figure*}
\begin{figure*}[!t]
	\centerline{\includegraphics[scale=0.5]{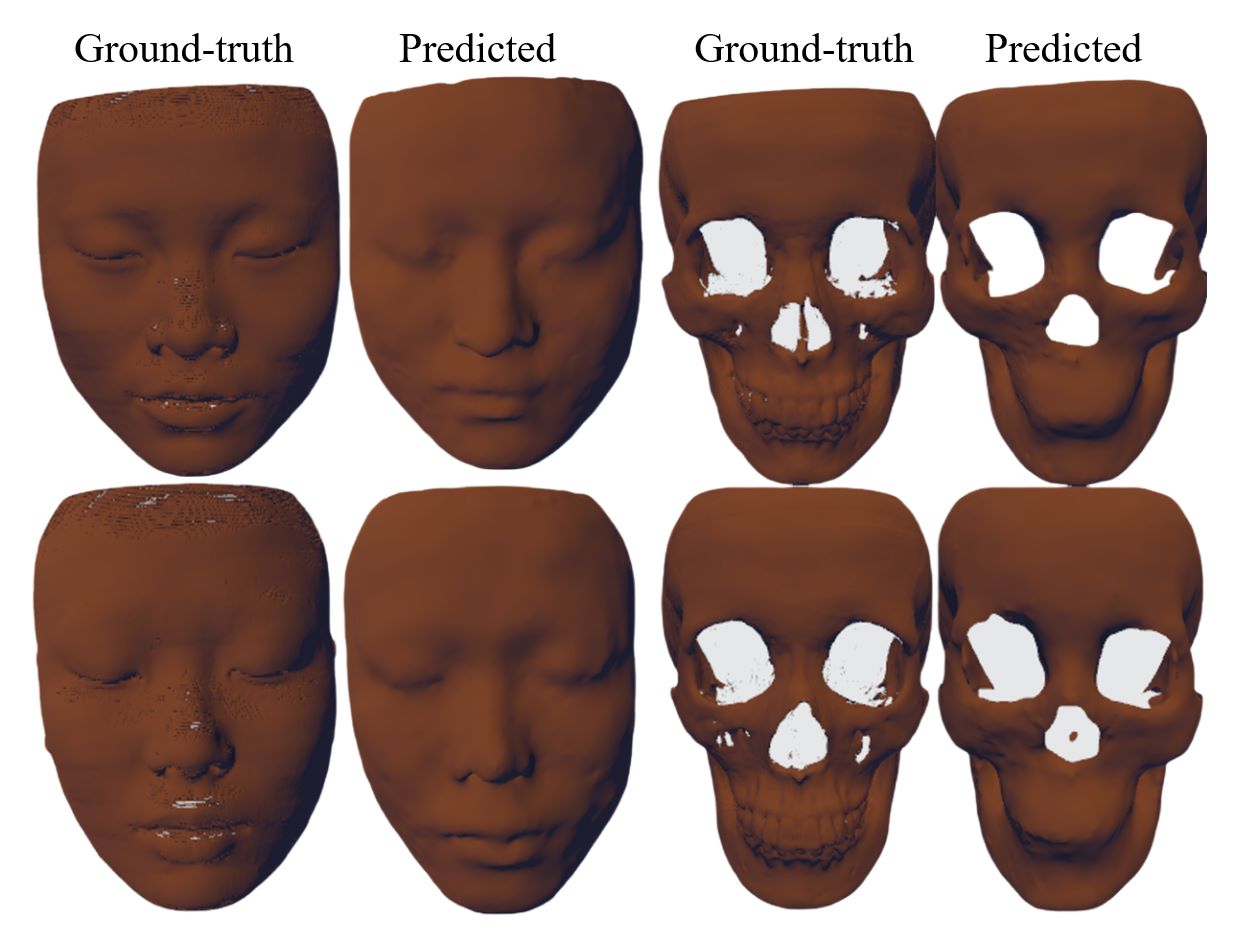}}
	\caption{Comparison between ground-truth and predicted meshes from two persons in the test set.}
	\label{fig11}
\end{figure*}
The results of the ablation experiment concerning the loss functions is shown in Table~\ref{tab3}.
After discarding $\mathcal{L}_{EMD}$, the performance of the network significantly degraded, reflecting the global correspondence improvement provided by one-to-one neighbor point matching.
Although abandoning the density distribution loss ($\mathcal{L}_{local}$) could improve the local region transformation, the global correspondence performance was obviously degraded.
This finding indicates that the loss can induce a more uniform point cloud distribution.
With the addition of auxiliary information, the global correspondence performance achieved for matching point clouds was slightly degraded, but this addition significantly improved the accuracy achieved in the local region.
The reason for this finding may be that our annotation scheme is based on prompts, and we consider that an intensively detailed annotation may improve the global transformation performance of the model by facilitating local correspondences.
The comparison results produced in local areas are shown in Fig. \ref{fig6} and Fig.~\ref{fig7}.
After adding the local loss, the spatial alignment of the ROI between the predicted and ground-truth point clouds is significantly improved, especially at the lip area.
This indicates that our proposed local loss effectively enhances point-to-point alignment in the local structure of the point cloud.
\subsubsection{Ablation Experiments}
Ablation experiments were performed by separately removing the added LIA-Net, GRU and relative position modules, and the results of our proposed method are shown in Table~\ref{tab4}.
The evaluation metrics significantly decreased, especially for the EMD, when the relative position module was removed.
This finding indicates that globally modeling orientation and relative position features can effectively improve the local and global performance of point cloud transformations.
Additionally, LIA-Net substantially increased its deformation accuracy, and this finding demonstrates the ability of the network to model local geometric structures.
An experiment involving the GRU module showed that the movement path in the first stage could guide the point-to-point movement in the second stage, thus realizing coarse-to-fine displacements.

The visualization results of coarse and fine predicted point clouds are shown in Fig.~\ref{fig8}. 
As observed, compared to the fine predictions, the coarse predictions exhibit many uneven distributions in local regions, increased outlier points in edge, and inferior detail representation in complex structures, such as coronoid process. 
This finding indicates our proposed LIA-Net can enhance global prediction accuracy by establishing local point-to-point correspondences and reducing outliers, thereby improving the overall prediction quality.
\subsubsection{Mesh Generation}
After generating the predicted point cloud, the data postprocessing method was employed to restructure the mesh.
We calculated the mean Hausdorff distance between the reconstructed mesh and the original ground-truth mesh from test set. 
The errors of reconstructed bone and face were 1.64$\pm$0.78 mm and 1.61$\pm$0.43 mm, respectively, which are acceptable in clinic. 
This finding indicates that a simple postprocessing-based reconstruction algorithm is sufficient for reconstructing highly accurate meshes for generating dense, precise point clouds in our framework.
Notably, mesh reconstruction was performed as a unified batch process in this paper, and manual fine-tuning the mesh by a professional engineer would likely yield better results in practical applications. 
The error visualizations of reconstructed meshes are shown in Fig.~\ref{fig9}.
The predicted mesh errors are mostly distributed within 2 mm. 
Among them, the errors at the organ edges are more noticeable, for example orbital, mandibular, and forehead, which is an inherent issue of the traditional reconstruction method. 
The prediction results of our method on one data sample are shown in Fig.~\ref{fig10}.
And the comparative results between ground-truth and predicted meshes are shown in Fig.~\ref{fig11}.
\section{Discussion}
In this paper, we propose a framework for bidirectionally transforming a face-bone mesh; the framework consists of a core point cloud transformation network and simple preprocessing and postprocessing modules.
Compared with the SOTA P2P-Conv approach~\cite{malei}, our Transformer-based network can process dense point clouds while achieving greater global correspondences and thus better accuracy and fewer outliers.
Our proposed LIA-Net module can model local point cloud features to compensate for the impaired local features caused by the Transformer-based network in the first stage, and it can be simply migrated to other existing frameworks.
More importantly, this framework does not require complex nonrigid registration operations between the input point cloud and templates in the preprocessing and postprocessing modules.
Furthermore, TCFNet can be used in other fields, such as point cloud completion and reconstruction, along with bidirectional transformations.

Although our approach achieved SOTA performance in bidirectional face-bone transformation tasks, two unsatisfactory features remain.
Inspired by deformable medical image registration, we initially found that labeling information based on prompts could obviously improve the transformation accuracy attained for local areas; however, the labeled bounding boxes were sparse, resulting in slightly degraded global transformation performance.
Although our method yielded relatively high accuracy, the reconstructed facial mesh using traditional reconstruction techniques from the predicted point cloud was subject to local distortions (such as concavities and unevenness), especially in dense point cloud regions (such as the nose). 
In the future, to overcome inherent limitations of traditional reconstruction methods, we will consider the neural implications of the surface reconstruction method~\cite{chen2024neural} to obtain smooth and continuous facial meshes. 
Furthermore, the dense segmentation information~\cite{zhangrunshi} and extensive prompts (for example, landmarks~\cite{zhanglandmark}) from our previous work may be employed to improve the quality of the reconstructed mesh.
\section{Conclusion}
In summary, we propose a simple and end-to-end framework, TCFNet, which is a Transformer-based coarse-to-fine point movement network, to simulate the relationships between bones and facial appearances during orthognathic surgery planning.
TCFNet can accurately and efficiently process dense point clouds, utilize global-to-local features derived from coarse-to-fine point movement, and overcome the traditional tradeoff between global and local matching.
Additionally, our proposed auxiliary information loss can leverage prior expert knowledge to achieve better local region matching effects.
Finally, our method produces optimal results in comparison with those of other SOTA methods.

\printcredits

\bibliographystyle{cas-model2-names}

\bibliography{cas-refs}

\begin{thebibliography}{55}
\expandafter\ifx\csname natexlab\endcsname\relax\def\natexlab#1{#1}\fi
\providecommand{\url}[1]{\texttt{#1}}
\providecommand{\href}[2]{#2}
\providecommand{\path}[1]{#1}
\providecommand{\DOIprefix}{doi:}
\providecommand{\ArXivprefix}{arXiv:}
\providecommand{\URLprefix}{URL: }
\providecommand{\Pubmedprefix}{pmid:}
\providecommand{\doi}[1]{\href{http://dx.doi.org/#1}{\path{#1}}}
\providecommand{\Pubmed}[1]{\href{pmid:#1}{\path{#1}}}
\providecommand{\bibinfo}[2]{#2}
\ifx\xfnm\relax \def\xfnm[#1]{\unskip,\space#1}\fi
\bibitem[{Achlioptas et~al.(2018)Achlioptas, Diamanti, Mitliagkas and
  Guibas}]{achlioptas2018learning}
\bibinfo{author}{Achlioptas, P.}, \bibinfo{author}{Diamanti, O.},
  \bibinfo{author}{Mitliagkas, I.}, \bibinfo{author}{Guibas, L.},
  \bibinfo{year}{2018}.
\newblock \bibinfo{title}{Learning representations and generative models for 3d
  point clouds}, in: \bibinfo{booktitle}{International conference on machine
  learning}, \bibinfo{organization}{PMLR}. pp. \bibinfo{pages}{40--49}.
\bibitem[{Bao et~al.(2024a)Bao, Zhang, Xiang, Liu, Cheng, Yang, Huang, Xiang,
  Cui, Lai, Huang, Wang, Qian and Yu}]{JBao}
\bibinfo{author}{Bao, J.}, \bibinfo{author}{Zhang, X.}, \bibinfo{author}{Xiang,
  S.}, \bibinfo{author}{Liu, H.}, \bibinfo{author}{Cheng, M.},
  \bibinfo{author}{Yang, Y.}, \bibinfo{author}{Huang, X.},
  \bibinfo{author}{Xiang, W.}, \bibinfo{author}{Cui, W.}, \bibinfo{author}{Lai,
  H.}, \bibinfo{author}{Huang, S.}, \bibinfo{author}{Wang, Y.},
  \bibinfo{author}{Qian, D.}, \bibinfo{author}{Yu, H.}, \bibinfo{year}{2024}a.
\newblock \bibinfo{title}{Deep learning–based facial and skeletal
  transformations for surgical planning}.
\newblock \bibinfo{journal}{Journal of Dental Research} \bibinfo{volume}{103},
  \bibinfo{pages}{809--819}.
\newblock \URLprefix \url{https://doi.org/10.1177/00220345241253186},
  \DOIprefix\doi{10.1177/00220345241253186},
  \href{http://arxiv.org/abs/https://doi.org/10.1177/00220345241253186}{\tt
  arXiv:https://doi.org/10.1177/00220345241253186}. \bibinfo{note}{pMID:
  38808566}.
\bibitem[{Bao et~al.(2024b)Bao, Zhang, Xiang, Liu, Cheng, Yang, Huang, Xiang,
  Cui, Lai, Huang, Wang, Qian and Yu}]{BaoJ}
\bibinfo{author}{Bao, J.}, \bibinfo{author}{Zhang, X.}, \bibinfo{author}{Xiang,
  S.}, \bibinfo{author}{Liu, H.}, \bibinfo{author}{Cheng, M.},
  \bibinfo{author}{Yang, Y.}, \bibinfo{author}{Huang, X.},
  \bibinfo{author}{Xiang, W.}, \bibinfo{author}{Cui, W.}, \bibinfo{author}{Lai,
  H.C.}, \bibinfo{author}{Huang, S.}, \bibinfo{author}{Wang, Y.},
  \bibinfo{author}{Qian, D.}, \bibinfo{author}{Yu, H.}, \bibinfo{year}{2024}b.
\newblock \bibinfo{title}{Deep learning-based facial and skeletal
  transformations for surgical planning}.
\newblock \bibinfo{journal}{JOURNAL OF DENTAL RESEARCH}
  \DOIprefix\doi{10.1177/00220345241253186}.
\bibitem[{Bianchi et~al.(2023)Bianchi, Seidita, Badiali, Lusetti, Saporosi,
  Pironi, Marchetti and Crimi}]{bianchi2023beauty}
\bibinfo{author}{Bianchi, A.}, \bibinfo{author}{Seidita, F.},
  \bibinfo{author}{Badiali, G.}, \bibinfo{author}{Lusetti, L.},
  \bibinfo{author}{Saporosi, C.}, \bibinfo{author}{Pironi, M.},
  \bibinfo{author}{Marchetti, C.}, \bibinfo{author}{Crimi, S.},
  \bibinfo{year}{2023}.
\newblock \bibinfo{title}{Is beauty a matter of volume distribution? proposal
  of a new aesthetic three-dimensional guide in orthognathic surgery}.
\newblock \bibinfo{journal}{Journal of Personalized Medicine}
  \bibinfo{volume}{13}, \bibinfo{pages}{936}.
\bibitem[{Chen et~al.(2024)Chen, Kumaralingam, Zhang, Song, Zhang, Zhang, Pham,
  Punithakumar, Lou, Zhang et~al.}]{chen2024neural}
\bibinfo{author}{Chen, H.}, \bibinfo{author}{Kumaralingam, L.},
  \bibinfo{author}{Zhang, S.}, \bibinfo{author}{Song, S.},
  \bibinfo{author}{Zhang, F.}, \bibinfo{author}{Zhang, H.},
  \bibinfo{author}{Pham, T.T.}, \bibinfo{author}{Punithakumar, K.},
  \bibinfo{author}{Lou, E.H.}, \bibinfo{author}{Zhang, Y.}, et~al.,
  \bibinfo{year}{2024}.
\newblock \bibinfo{title}{Neural implicit surface reconstruction of freehand 3d
  ultrasound volume with geometric constraints}.
\newblock \bibinfo{journal}{Medical Image Analysis} \bibinfo{volume}{98},
  \bibinfo{pages}{103305}.
\bibitem[{Chen et~al.(2022)Chen, Zhu, Chen and Yu}]{chen2022efficient}
\bibinfo{author}{Chen, W.}, \bibinfo{author}{Zhu, X.}, \bibinfo{author}{Chen,
  G.}, \bibinfo{author}{Yu, B.}, \bibinfo{year}{2022}.
\newblock \bibinfo{title}{Efficient point cloud analysis using hilbert curve},
  in: \bibinfo{booktitle}{European Conference on Computer Vision},
  \bibinfo{organization}{Springer}. pp. \bibinfo{pages}{730--747}.
\bibitem[{Denadai et~al.(2020)Denadai, Pai and Lo}]{RafaelDenadai}
\bibinfo{author}{Denadai, R.}, \bibinfo{author}{Pai, B.C.},
  \bibinfo{author}{Lo, L.J.}, \bibinfo{year}{2020}.
\newblock \bibinfo{title}{Balancing the dental occlusion and facial aesthetic
  features in cleft orthognathic surgery: Patient-centered concept for
  computer-aided planning}.
\newblock \bibinfo{journal}{Biomedical Journal} \bibinfo{volume}{43},
  \bibinfo{pages}{143--145}.
\newblock \URLprefix
  \url{https://www.sciencedirect.com/science/article/pii/S231941701930530X},
  \DOIprefix\doi{https://doi.org/10.1016/j.bj.2019.12.008}.
\bibitem[{Elnagar et~al.(2020)Elnagar, Aronovich and Kusnoto}]{Mohammed}
\bibinfo{author}{Elnagar, M.H.}, \bibinfo{author}{Aronovich, S.},
  \bibinfo{author}{Kusnoto, B.}, \bibinfo{year}{2020}.
\newblock \bibinfo{title}{Digital workflow for combined orthodontics and
  orthognathic surgery}.
\newblock \bibinfo{journal}{Oral and Maxillofacial Surgery Clinics of North
  America} \bibinfo{volume}{32}, \bibinfo{pages}{1--14}.
\newblock \URLprefix
  \url{https://www.sciencedirect.com/science/article/pii/S1042369919300688},
  \DOIprefix\doi{https://doi.org/10.1016/j.coms.2019.08.004}.
  \bibinfo{note}{orthodontics for the Oral and Maxillofacial Surgery Patient}.
\bibitem[{Fang et~al.(2024)Fang, Kim, Xu, Kuang, Lampen, Lee, Deng, Liebschner,
  Xia, Gateno and Yan}]{xifang}
\bibinfo{author}{Fang, X.}, \bibinfo{author}{Kim, D.}, \bibinfo{author}{Xu,
  X.}, \bibinfo{author}{Kuang, T.}, \bibinfo{author}{Lampen, N.},
  \bibinfo{author}{Lee, J.}, \bibinfo{author}{Deng, H.H.},
  \bibinfo{author}{Liebschner, M.A.}, \bibinfo{author}{Xia, J.J.},
  \bibinfo{author}{Gateno, J.}, \bibinfo{author}{Yan, P.},
  \bibinfo{year}{2024}.
\newblock \bibinfo{title}{Correspondence attention for facial appearance
  simulation}.
\newblock \bibinfo{journal}{Medical Image Analysis} \bibinfo{volume}{93},
  \bibinfo{pages}{103094}.
\newblock \URLprefix
  \url{https://www.sciencedirect.com/science/article/pii/S1361841524000197},
  \DOIprefix\doi{https://doi.org/10.1016/j.media.2024.103094}.
\bibitem[{Graham et~al.(2018)Graham, Engelcke and Maaten}]{GrahamBenjamin}
\bibinfo{author}{Graham, B.}, \bibinfo{author}{Engelcke, M.},
  \bibinfo{author}{Maaten, L.v.d.}, \bibinfo{year}{2018}.
\newblock \bibinfo{title}{3d semantic segmentation with submanifold sparse
  convolutional networks}, in: \bibinfo{booktitle}{2018 IEEE/CVF Conference on
  Computer Vision and Pattern Recognition}, pp. \bibinfo{pages}{9224--9232}.
\newblock \DOIprefix\doi{10.1109/CVPR.2018.00961}.
\bibitem[{Guo et~al.(2021)Guo, Cai, Liu, Mu, Martin and Hu}]{GuoMengHaoPCT}
\bibinfo{author}{Guo, M.H.}, \bibinfo{author}{Cai, J.X.}, \bibinfo{author}{Liu,
  Z.N.}, \bibinfo{author}{Mu, T.J.}, \bibinfo{author}{Martin, R.R.},
  \bibinfo{author}{Hu, S.M.}, \bibinfo{year}{2021}.
\newblock \bibinfo{title}{Pct: Point cloud transformer}.
\newblock \bibinfo{journal}{COMPUTATIONAL VISUAL MEDIA} \bibinfo{volume}{7},
  \bibinfo{pages}{187--199}.
\newblock \DOIprefix\doi{10.1007/s41095-021-0229-5}.
\bibitem[{Han et~al.(2021)Han, Uneri, Vijayan, Wu, Vagdargi, Sheth, Vogt,
  Kleinszig, Osgood and Siewerdsen}]{rhan}
\bibinfo{author}{Han, R.}, \bibinfo{author}{Uneri, A.},
  \bibinfo{author}{Vijayan, R.}, \bibinfo{author}{Wu, P.},
  \bibinfo{author}{Vagdargi, P.}, \bibinfo{author}{Sheth, N.},
  \bibinfo{author}{Vogt, S.}, \bibinfo{author}{Kleinszig, G.},
  \bibinfo{author}{Osgood, G.}, \bibinfo{author}{Siewerdsen, J.},
  \bibinfo{year}{2021}.
\newblock \bibinfo{title}{Fracture reduction planning and guidance in
  orthopaedic trauma surgery via multi-body image registration}.
\newblock \bibinfo{journal}{Medical Image Analysis} \bibinfo{volume}{68},
  \bibinfo{pages}{101917}.
\newblock \URLprefix
  \url{https://www.sciencedirect.com/science/article/pii/S1361841520302814},
  \DOIprefix\doi{https://doi.org/10.1016/j.media.2020.101917}.
\bibitem[{Huang et~al.(2023)Huang, Ding, Zhang, Tai, Zhang, Chen, Wang and
  Liu}]{huang2023learning}
\bibinfo{author}{Huang, T.}, \bibinfo{author}{Ding, Z.},
  \bibinfo{author}{Zhang, J.}, \bibinfo{author}{Tai, Y.},
  \bibinfo{author}{Zhang, Z.}, \bibinfo{author}{Chen, M.},
  \bibinfo{author}{Wang, C.}, \bibinfo{author}{Liu, Y.}, \bibinfo{year}{2023}.
\newblock \bibinfo{title}{Learning to measure the point cloud reconstruction
  loss in a representation space}, in: \bibinfo{booktitle}{Proceedings of the
  IEEE/CVF Conference on Computer Vision and Pattern Recognition}, pp.
  \bibinfo{pages}{12208--12217}.
\bibitem[{Huang et~al.(2022)Huang, Yang, Zhang, Cui, Zou, Chen, Zhao and
  Liu}]{huang2022learning}
\bibinfo{author}{Huang, T.}, \bibinfo{author}{Yang, X.},
  \bibinfo{author}{Zhang, J.}, \bibinfo{author}{Cui, J.}, \bibinfo{author}{Zou,
  H.}, \bibinfo{author}{Chen, J.}, \bibinfo{author}{Zhao, X.},
  \bibinfo{author}{Liu, Y.}, \bibinfo{year}{2022}.
\newblock \bibinfo{title}{Learning to train a point cloud reconstruction
  network without matching}, in: \bibinfo{booktitle}{European Conference on
  Computer Vision}, \bibinfo{organization}{Springer}. pp.
  \bibinfo{pages}{179--194}.
\bibitem[{Jia et~al.(2021)Jia, Zhao, Xin, Duan, Pan, Wu, Li and Zhou}]{BinJia}
\bibinfo{author}{Jia, B.}, \bibinfo{author}{Zhao, J.}, \bibinfo{author}{Xin,
  S.}, \bibinfo{author}{Duan, F.}, \bibinfo{author}{Pan, Z.},
  \bibinfo{author}{Wu, Z.}, \bibinfo{author}{Li, J.}, \bibinfo{author}{Zhou,
  M.}, \bibinfo{year}{2021}.
\newblock \bibinfo{title}{Craniofacial reconstruction based on heat flow
  geodesic grid regression (hf-ggr) model}.
\newblock \bibinfo{journal}{Computers and Graphics} \bibinfo{volume}{97},
  \bibinfo{pages}{258--267}.
\bibitem[{Kazhdan et~al.(2006)Kazhdan, Bolitho and Hoppe}]{kazhdan2006poisson}
\bibinfo{author}{Kazhdan, M.}, \bibinfo{author}{Bolitho, M.},
  \bibinfo{author}{Hoppe, H.}, \bibinfo{year}{2006}.
\newblock \bibinfo{title}{Poisson surface reconstruction}, in:
  \bibinfo{booktitle}{Proceedings of the fourth Eurographics symposium on
  Geometry processing}.
\bibitem[{Kim et~al.(2019)Kim, Kuang, Rodrigues, Gateno, Shen, Wang, Deng,
  Yuan, Alfi, Liebschner and Xia}]{KimDaeseung_miccai}
\bibinfo{author}{Kim, D.}, \bibinfo{author}{Kuang, T.},
  \bibinfo{author}{Rodrigues, Y.L.}, \bibinfo{author}{Gateno, J.},
  \bibinfo{author}{Shen, S.G.F.}, \bibinfo{author}{Wang, X.},
  \bibinfo{author}{Deng, H.}, \bibinfo{author}{Yuan, P.},
  \bibinfo{author}{Alfi, D.M.}, \bibinfo{author}{Liebschner, M.A.K.},
  \bibinfo{author}{Xia, J.J.}, \bibinfo{year}{2019}.
\newblock \bibinfo{title}{A new approach of predicting facial changes following
  orthognathic surgery using realistic lip sliding effect}, in:
  \bibinfo{editor}{Shen, D.}, \bibinfo{editor}{Liu, T.},
  \bibinfo{editor}{Peters, T.}, \bibinfo{editor}{Staib, L.},
  \bibinfo{editor}{Essert, C.}, \bibinfo{editor}{Zhou, S.},
  \bibinfo{editor}{Yap, P.}, \bibinfo{editor}{Khan, A.} (Eds.),
  \bibinfo{booktitle}{MEDICAL IMAGE COMPUTING AND COMPUTER ASSISTED
  INTERVENTION - MICCAI 2019, PT V}, pp. \bibinfo{pages}{336--344}.
\newblock \DOIprefix\doi{10.1007/978-3-030-32254-0\_38}. \bibinfo{note}{10th
  International Workshop on Machine Learning in Medical Imaging (MLMI) / 22nd
  International Conference on Medical Image Computing and Computer-Assisted
  Intervention (MICCAI), Shenzhen, PEOPLES R CHINA, OCT 13-17, 2019}.
\bibitem[{Kim et~al.(2021)Kim, Kuang, Rodrigues, Gateno, Shen, Wang, Stein,
  Deng, Liebschner and Xia}]{KimDaeseung_mia}
\bibinfo{author}{Kim, D.}, \bibinfo{author}{Kuang, T.},
  \bibinfo{author}{Rodrigues, Y.L.}, \bibinfo{author}{Gateno, J.},
  \bibinfo{author}{Shen, S.G.F.}, \bibinfo{author}{Wang, X.},
  \bibinfo{author}{Stein, K.}, \bibinfo{author}{Deng, H.H.},
  \bibinfo{author}{Liebschner, M.A.K.}, \bibinfo{author}{Xia, J.J.},
  \bibinfo{year}{2021}.
\newblock \bibinfo{title}{A novel incremental simulation of facial changes
  following orthognathic surgery using fem with realistic lip sliding effect}.
\newblock \bibinfo{journal}{MEDICAL IMAGE ANALYSIS} \bibinfo{volume}{72}.
\newblock \DOIprefix\doi{10.1016/j.media.2021.102095}.
\bibitem[{Lampen et~al.(2023)Lampen, Kim, Xu, Fang, Lee, Kuang, Deng,
  Liebschner, Xia, Gateno and Yan}]{LampenNathan}
\bibinfo{author}{Lampen, N.}, \bibinfo{author}{Kim, D.}, \bibinfo{author}{Xu,
  X.}, \bibinfo{author}{Fang, X.}, \bibinfo{author}{Lee, J.},
  \bibinfo{author}{Kuang, T.}, \bibinfo{author}{Deng, H.H.},
  \bibinfo{author}{Liebschner, M.A.K.}, \bibinfo{author}{Xia, J.J.},
  \bibinfo{author}{Gateno, J.}, \bibinfo{author}{Yan, P.},
  \bibinfo{year}{2023}.
\newblock \bibinfo{title}{Spatiotemporal incremental mechanics modeling
  of facial tissue change}, in: \bibinfo{editor}{Greenspan, H.},
  \bibinfo{editor}{Madabhushi, A.}, \bibinfo{editor}{Mousavi, P.},
  \bibinfo{editor}{Salcudean, S.}, \bibinfo{editor}{Duncan, J.},
  \bibinfo{editor}{Syeda-Mahmood, T.}, \bibinfo{editor}{Taylor, R.} (Eds.),
  \bibinfo{booktitle}{Medical Image Computing and Computer Assisted
  Intervention -- MICCAI 2023}, \bibinfo{publisher}{Springer Nature
  Switzerland}, \bibinfo{address}{Cham}. pp. \bibinfo{pages}{566--575}.
\bibitem[{Lan et~al.(2019)Lan, Yu, Yu and Davis}]{LanShiyi}
\bibinfo{author}{Lan, S.}, \bibinfo{author}{Yu, R.}, \bibinfo{author}{Yu, G.},
  \bibinfo{author}{Davis, L.S.}, \bibinfo{year}{2019}.
\newblock \bibinfo{title}{Modeling local geometric structure of 3d point clouds
  using geo-cnn}, in: \bibinfo{booktitle}{Proceedings of the IEEE/cvf
  conference on computer vision and pattern recognition}, pp.
  \bibinfo{pages}{998--1008}.
\bibitem[{Liu et~al.(2021)Liu, Lin, Cao, Hu, Wei, Zhang, Lin and
  Guo}]{liu2021swin}
\bibinfo{author}{Liu, Z.}, \bibinfo{author}{Lin, Y.}, \bibinfo{author}{Cao,
  Y.}, \bibinfo{author}{Hu, H.}, \bibinfo{author}{Wei, Y.},
  \bibinfo{author}{Zhang, Z.}, \bibinfo{author}{Lin, S.}, \bibinfo{author}{Guo,
  B.}, \bibinfo{year}{2021}.
\newblock \bibinfo{title}{Swin transformer: Hierarchical vision transformer
  using shifted windows}, in: \bibinfo{booktitle}{Proceedings of the IEEE/CVF
  international conference on computer vision}, pp.
  \bibinfo{pages}{10012--10022}.
\bibitem[{Liu et~al.(2023)Liu, Yang, Tang, Yang and Han}]{liu2023flatformer}
\bibinfo{author}{Liu, Z.}, \bibinfo{author}{Yang, X.}, \bibinfo{author}{Tang,
  H.}, \bibinfo{author}{Yang, S.}, \bibinfo{author}{Han, S.},
  \bibinfo{year}{2023}.
\newblock \bibinfo{title}{Flatformer: Flattened window attention for efficient
  point cloud transformer}, in: \bibinfo{booktitle}{Proceedings of the IEEE/CVF
  Conference on Computer Vision and Pattern Recognition}, pp.
  \bibinfo{pages}{1200--1211}.
\bibitem[{Luo et~al.(2024)Luo, Du, Zhu, Yu, Fu and Han}]{LuoZhongjin}
\bibinfo{author}{Luo, Z.}, \bibinfo{author}{Du, D.}, \bibinfo{author}{Zhu, H.},
  \bibinfo{author}{Yu, Y.}, \bibinfo{author}{Fu, H.}, \bibinfo{author}{Han,
  X.}, \bibinfo{year}{2024}.
\newblock \bibinfo{title}{Sketchmetaface: A learning-based sketching interface
  for high-fidelity 3d character face modeling}.
\newblock \bibinfo{journal}{IEEE Transactions on Visualization and Computer
  Graphics} \bibinfo{volume}{30}, \bibinfo{pages}{5260--5275}.
\newblock \DOIprefix\doi{10.1109/TVCG.2023.3291703}.
\bibitem[{Ma et~al.(2023a)Ma, Lian, Kim, Xiao, Wei, Liu, Kuang, Ghanbari, Li,
  Gateno, Shen, Wang, Shen, Xia and Yap}]{malei}
\bibinfo{author}{Ma, L.}, \bibinfo{author}{Lian, C.}, \bibinfo{author}{Kim,
  D.}, \bibinfo{author}{Xiao, D.}, \bibinfo{author}{Wei, D.},
  \bibinfo{author}{Liu, Q.}, \bibinfo{author}{Kuang, T.},
  \bibinfo{author}{Ghanbari, M.}, \bibinfo{author}{Li, G.},
  \bibinfo{author}{Gateno, J.}, \bibinfo{author}{Shen, S.G.},
  \bibinfo{author}{Wang, L.}, \bibinfo{author}{Shen, D.}, \bibinfo{author}{Xia,
  J.J.}, \bibinfo{author}{Yap, P.T.}, \bibinfo{year}{2023}a.
\newblock \bibinfo{title}{Bidirectional prediction of facial and bony shapes
  for orthognathic surgical planning}.
\newblock \bibinfo{journal}{Medical Image Analysis} \bibinfo{volume}{83},
  \bibinfo{pages}{102644}.
\newblock \URLprefix
  \url{https://www.sciencedirect.com/science/article/pii/S1361841522002729},
  \DOIprefix\doi{https://doi.org/10.1016/j.media.2022.102644}.
\bibitem[{Ma et~al.(2023b)Ma, Xiao, Kim, Lian, Kuang, Liu, Deng, Yang,
  Liebschner, Gateno, Xia and Yap}]{maleitmi}
\bibinfo{author}{Ma, L.}, \bibinfo{author}{Xiao, D.}, \bibinfo{author}{Kim,
  D.}, \bibinfo{author}{Lian, C.}, \bibinfo{author}{Kuang, T.},
  \bibinfo{author}{Liu, Q.}, \bibinfo{author}{Deng, H.}, \bibinfo{author}{Yang,
  E.}, \bibinfo{author}{Liebschner, M.A.K.}, \bibinfo{author}{Gateno, J.},
  \bibinfo{author}{Xia, J.J.}, \bibinfo{author}{Yap, P.T.},
  \bibinfo{year}{2023}b.
\newblock \bibinfo{title}{Simulation of postoperative facial appearances via
  geometric deep learning for efficient orthognathic surgical planning}.
\newblock \bibinfo{journal}{IEEE Transactions on Medical Imaging}
  \bibinfo{volume}{42}, \bibinfo{pages}{336--345}.
\newblock \DOIprefix\doi{10.1109/TMI.2022.3180078}.
\bibitem[{Millesi et~al.(2023)Millesi, Zimmermann and
  Eltz}]{millesi2023surgery}
\bibinfo{author}{Millesi, G.A.}, \bibinfo{author}{Zimmermann, M.},
  \bibinfo{author}{Eltz, M.}, \bibinfo{year}{2023}.
\newblock \bibinfo{title}{Surgery first and surgery early treatment approach in
  orthognathic surgery}.
\newblock \bibinfo{journal}{Oral and Maxillofacial Surgery Clinics}
  \bibinfo{volume}{35}, \bibinfo{pages}{71--82}.
\bibitem[{Mollemans et~al.(2007)Mollemans, Schutyser, Nadjmi, Maes and
  Suetens}]{MollemansW}
\bibinfo{author}{Mollemans, W.}, \bibinfo{author}{Schutyser, F.},
  \bibinfo{author}{Nadjmi, N.}, \bibinfo{author}{Maes, F.},
  \bibinfo{author}{Suetens, P.}, \bibinfo{year}{2007}.
\newblock \bibinfo{title}{Predicting soft tissue deformations for a
  maxillofacial surgery planning system: From computational strategies to a
  complete clinical validation}.
\newblock \bibinfo{journal}{MEDICAL IMAGE ANALYSIS} \bibinfo{volume}{11},
  \bibinfo{pages}{282--301}.
\newblock \DOIprefix\doi{10.1016/j.media.2007.02.003}.
\bibitem[{Paysan et~al.(2009)Paysan, L{\"u}thi, Albrecht, Lerch, Amberg,
  Santini and Vetter}]{PaysanPascal}
\bibinfo{author}{Paysan, P.}, \bibinfo{author}{L{\"u}thi, M.},
  \bibinfo{author}{Albrecht, T.}, \bibinfo{author}{Lerch, A.},
  \bibinfo{author}{Amberg, B.}, \bibinfo{author}{Santini, F.},
  \bibinfo{author}{Vetter, T.}, \bibinfo{year}{2009}.
\newblock \bibinfo{title}{Face reconstruction from skull shapes and physical
  attributes}, in: \bibinfo{booktitle}{Pattern Recognition: 31st DAGM
  Symposium, Jena, Germany, September 9-11, 2009. Proceedings 31},
  \bibinfo{organization}{Springer}. pp. \bibinfo{pages}{232--241}.
\bibitem[{Peng et~al.(2024)Peng, Wu, Jiang, Chen, Zhao, Tian and Jia}]{OACNNs}
\bibinfo{author}{Peng, B.}, \bibinfo{author}{Wu, X.}, \bibinfo{author}{Jiang,
  L.}, \bibinfo{author}{Chen, Y.}, \bibinfo{author}{Zhao, H.},
  \bibinfo{author}{Tian, Z.}, \bibinfo{author}{Jia, J.}, \bibinfo{year}{2024}.
\newblock \bibinfo{title}{Oa-cnns: Omni-adaptive sparse cnns for 3d semantic
  segmentation}, in: \bibinfo{booktitle}{Proceedings of the IEEE/CVF Conference
  on Computer Vision and Pattern Recognition}, pp.
  \bibinfo{pages}{21305--21315}.
\bibitem[{Qi et~al.(2017a)Qi, Su, Mo and Guibas}]{QiCharlesR}
\bibinfo{author}{Qi, C.R.}, \bibinfo{author}{Su, H.}, \bibinfo{author}{Mo, K.},
  \bibinfo{author}{Guibas, L.J.}, \bibinfo{year}{2017}a.
\newblock \bibinfo{title}{Pointnet: Deep learning on point sets for 3d
  classification and segmentation}, in: \bibinfo{booktitle}{Proceedings of the
  IEEE conference on computer vision and pattern recognition}, pp.
  \bibinfo{pages}{652--660}.
\bibitem[{Qi et~al.(2017b)Qi, Yi, Su and Guibas}]{QiCharlesRplus}
\bibinfo{author}{Qi, C.R.}, \bibinfo{author}{Yi, L.}, \bibinfo{author}{Su, H.},
  \bibinfo{author}{Guibas, L.J.}, \bibinfo{year}{2017}b.
\newblock \bibinfo{title}{Pointnet plus plus : Deep hierarchical feature
  learning on point sets in a metric space}, in: \bibinfo{editor}{Guyon, I.},
  \bibinfo{editor}{Luxburg, U.}, \bibinfo{editor}{Bengio, S.},
  \bibinfo{editor}{Wallach, H.}, \bibinfo{editor}{Fergus, R.},
  \bibinfo{editor}{Vishwanathan, S.}, \bibinfo{editor}{Garnett, R.} (Eds.),
  \bibinfo{booktitle}{ADVANCES IN NEURAL INFORMATION PROCESSING SYSTEMS 30
  (NIPS 2017)}.
\newblock \bibinfo{note}{31st Annual Conference on Neural Information
  Processing Systems (NIPS), Long Beach, CA, DEC 04-09, 2017}.
\bibitem[{R et~al.(2021)R, Tewari, Seidel, Elgharib and
  Theobalt}]{RMallikarjunB}
\bibinfo{author}{R, M.B.}, \bibinfo{author}{Tewari, A.},
  \bibinfo{author}{Seidel, H.P.}, \bibinfo{author}{Elgharib, M.},
  \bibinfo{author}{Theobalt, C.}, \bibinfo{year}{2021}.
\newblock \bibinfo{title}{Learning complete 3d morphable face models from
  images and videos}, in: \bibinfo{booktitle}{2021 IEEE/CVF Conference on
  Computer Vision and Pattern Recognition (CVPR)}, pp.
  \bibinfo{pages}{3360--3370}.
\newblock \DOIprefix\doi{10.1109/CVPR46437.2021.00337}.
\bibitem[{Ruggiero et~al.(2023)Ruggiero, Borghi, Bevini, Badiali, Lunari,
  Dunaway and Marchetti}]{RuggieroFederica}
\bibinfo{author}{Ruggiero, F.}, \bibinfo{author}{Borghi, A.},
  \bibinfo{author}{Bevini, M.}, \bibinfo{author}{Badiali, G.},
  \bibinfo{author}{Lunari, O.}, \bibinfo{author}{Dunaway, D.},
  \bibinfo{author}{Marchetti, C.}, \bibinfo{year}{2023}.
\newblock \bibinfo{title}{Soft tissue prediction in orthognathic surgery:
  Improving accuracy by means of anatomical details}.
\newblock \bibinfo{journal}{PLOS ONE} \bibinfo{volume}{18}.
\newblock \DOIprefix\doi{10.1371/journal.pone.0294640}.
\bibitem[{Wang et~al.(2022)Wang, Chen, Yu, Ma, Li and Liu}]{WangLizhen}
\bibinfo{author}{Wang, L.}, \bibinfo{author}{Chen, Z.}, \bibinfo{author}{Yu,
  T.}, \bibinfo{author}{Ma, C.}, \bibinfo{author}{Li, L.},
  \bibinfo{author}{Liu, Y.}, \bibinfo{year}{2022}.
\newblock \bibinfo{title}{Faceverse: a fine-grained and detail-controllable 3d
  face morphable model from a hybrid dataset}, in: \bibinfo{booktitle}{2022
  IEEE/CVF Conference on Computer Vision and Pattern Recognition (CVPR)}, pp.
  \bibinfo{pages}{20301--20310}.
\newblock \DOIprefix\doi{10.1109/CVPR52688.2022.01969}.
\bibitem[{Wang(2023)}]{wang2023octformer}
\bibinfo{author}{Wang, P.S.}, \bibinfo{year}{2023}.
\newblock \bibinfo{title}{Octformer: Octree-based transformers for 3d point
  clouds}.
\newblock \bibinfo{journal}{ACM Transactions on Graphics (TOG)}
  \bibinfo{volume}{42}, \bibinfo{pages}{1--11}.
\bibitem[{Wang et~al.(2019)Wang, Sun, Liu, Sarma, Bronstein and
  Solomon}]{WangYue}
\bibinfo{author}{Wang, Y.}, \bibinfo{author}{Sun, Y.}, \bibinfo{author}{Liu,
  Z.}, \bibinfo{author}{Sarma, S.E.}, \bibinfo{author}{Bronstein, M.M.},
  \bibinfo{author}{Solomon, J.M.}, \bibinfo{year}{2019}.
\newblock \bibinfo{title}{Dynamic graph cnn for learning on point clouds}.
\newblock \bibinfo{journal}{ACM Transactions on Graphics (tog)}
  \bibinfo{volume}{38}, \bibinfo{pages}{1--12}.
\bibitem[{Wen et~al.(2021)Wen, Xiang, Han, Cao, Wan, Zheng and Liu}]{PMPNet}
\bibinfo{author}{Wen, X.}, \bibinfo{author}{Xiang, P.}, \bibinfo{author}{Han,
  Z.}, \bibinfo{author}{Cao, Y.P.}, \bibinfo{author}{Wan, P.},
  \bibinfo{author}{Zheng, W.}, \bibinfo{author}{Liu, Y.S.},
  \bibinfo{year}{2021}.
\newblock \bibinfo{title}{Pmp-net: Point cloud completion by learning
  multi-step point moving paths}, in: \bibinfo{booktitle}{2021 IEEE/CVF
  Conference on Computer Vision and Pattern Recognition (CVPR)}, pp.
  \bibinfo{pages}{7439--7448}.
\newblock \DOIprefix\doi{10.1109/CVPR46437.2021.00736}.
\bibitem[{Wen et~al.(2023)Wen, Xiang, Han, Cao, Wan, Zheng and
  Liu}]{PMPNetplus}
\bibinfo{author}{Wen, X.}, \bibinfo{author}{Xiang, P.}, \bibinfo{author}{Han,
  Z.}, \bibinfo{author}{Cao, Y.P.}, \bibinfo{author}{Wan, P.},
  \bibinfo{author}{Zheng, W.}, \bibinfo{author}{Liu, Y.S.},
  \bibinfo{year}{2023}.
\newblock \bibinfo{title}{Pmp-net++: Point cloud completion by
  transformer-enhanced multi-step point moving paths}.
\newblock \bibinfo{journal}{IEEE Transactions on Pattern Analysis and Machine
  Intelligence} \bibinfo{volume}{45}, \bibinfo{pages}{852--867}.
\newblock \DOIprefix\doi{10.1109/TPAMI.2022.3159003}.
\bibitem[{Wu et~al.(2021)Wu, Pan, Zhang, Wang, Liu and Lin}]{WuTong2021}
\bibinfo{author}{Wu, T.}, \bibinfo{author}{Pan, L.}, \bibinfo{author}{Zhang,
  J.}, \bibinfo{author}{Wang, T.}, \bibinfo{author}{Liu, Z.},
  \bibinfo{author}{Lin, D.}, \bibinfo{year}{2021}.
\newblock \bibinfo{title}{Density-aware chamfer distance as a comprehensive
  metric for point cloud completion}, in: \bibinfo{editor}{Ranzato, M.},
  \bibinfo{editor}{Beygelzimer, A.}, \bibinfo{editor}{Dauphin, Y.},
  \bibinfo{editor}{Liang, P.}, \bibinfo{editor}{Vaughan, J.} (Eds.),
  \bibinfo{booktitle}{ADVANCES IN NEURAL INFORMATION PROCESSING SYSTEMS 34
  (NEURIPS 2021)}.
\newblock \bibinfo{note}{35th Annual Conference on Neural Information
  Processing Systems (NeurIPS), ELECTR NETWORK, DEC 06-14, 2021}.
\bibitem[{Wu et~al.(2019)Wu, Qi and Fuxin}]{WuWenxuan}
\bibinfo{author}{Wu, W.}, \bibinfo{author}{Qi, Z.}, \bibinfo{author}{Fuxin,
  L.}, \bibinfo{year}{2019}.
\newblock \bibinfo{title}{Pointconv: Deep convolutional networks on 3d point
  clouds}, in: \bibinfo{booktitle}{2019 IEEE/CVF Conference on Computer Vision
  and Pattern Recognition (CVPR)}, pp. \bibinfo{pages}{9613--9622}.
\newblock \DOIprefix\doi{10.1109/CVPR.2019.00985}.
\bibitem[{Wu et~al.(2024)Wu, Jiang, Wang, Liu, Liu, Qiao, Ouyang, He and
  Zhao}]{WuXiaoyangv3}
\bibinfo{author}{Wu, X.}, \bibinfo{author}{Jiang, L.}, \bibinfo{author}{Wang,
  P.S.}, \bibinfo{author}{Liu, Z.}, \bibinfo{author}{Liu, X.},
  \bibinfo{author}{Qiao, Y.}, \bibinfo{author}{Ouyang, W.},
  \bibinfo{author}{He, T.}, \bibinfo{author}{Zhao, H.}, \bibinfo{year}{2024}.
\newblock \bibinfo{title}{Point transformer v3: Simpler faster stronger}, in:
  \bibinfo{booktitle}{Proceedings of the IEEE/CVF Conference on Computer Vision
  and Pattern Recognition}, pp. \bibinfo{pages}{4840--4851}.
\bibitem[{Wu et~al.(2022)Wu, Lao, Jiang, Liu and Zhao}]{WuXiaoyangv2}
\bibinfo{author}{Wu, X.}, \bibinfo{author}{Lao, Y.}, \bibinfo{author}{Jiang,
  L.}, \bibinfo{author}{Liu, X.}, \bibinfo{author}{Zhao, H.},
  \bibinfo{year}{2022}.
\newblock \bibinfo{title}{Point transformer v2: Grouped vector attention and
  partition-based pooling}.
\newblock \bibinfo{journal}{Advances in Neural Information Processing Systems}
  \bibinfo{volume}{35}, \bibinfo{pages}{33330--33342}.
\bibitem[{Xiao et~al.(2021)Xiao, Lian, Wang, Deng, Lin, Thung, Zhu, Yuan,
  Perez, Gateno, Shen, Yap, Xia and Shen}]{XiaoDeqiang}
\bibinfo{author}{Xiao, D.}, \bibinfo{author}{Lian, C.}, \bibinfo{author}{Wang,
  L.}, \bibinfo{author}{Deng, H.}, \bibinfo{author}{Lin, H.Y.},
  \bibinfo{author}{Thung, K.H.}, \bibinfo{author}{Zhu, J.},
  \bibinfo{author}{Yuan, P.}, \bibinfo{author}{Perez, L.},
  \bibinfo{author}{Gateno, J.}, \bibinfo{author}{Shen, S.G.},
  \bibinfo{author}{Yap, P.T.}, \bibinfo{author}{Xia, J.J.},
  \bibinfo{author}{Shen, D.}, \bibinfo{year}{2021}.
\newblock \bibinfo{title}{Estimating reference shape model for personalized
  surgical reconstruction of craniomaxillofacial defects}.
\newblock \bibinfo{journal}{IEEE Transactions on Biomedical Engineering}
  \bibinfo{volume}{68}, \bibinfo{pages}{362--373}.
\newblock \DOIprefix\doi{10.1109/TBME.2020.2990586}.
\bibitem[{Xiao et~al.(2019)Xiao, Wang, Deng, Thung, Zhu, Yuan, Rodrigues,
  Perez, Crecelius, Gateno, Kuang, Shen, Kim, Alfi, Yap, Xia and
  Shen}]{XiaoDeqianghuiyi}
\bibinfo{author}{Xiao, D.}, \bibinfo{author}{Wang, L.}, \bibinfo{author}{Deng,
  H.}, \bibinfo{author}{Thung, K.H.}, \bibinfo{author}{Zhu, J.},
  \bibinfo{author}{Yuan, P.}, \bibinfo{author}{Rodrigues, Y.L.},
  \bibinfo{author}{Perez, L.}, \bibinfo{author}{Crecelius, C.E.},
  \bibinfo{author}{Gateno, J.}, \bibinfo{author}{Kuang, T.},
  \bibinfo{author}{Shen, S.G.F.}, \bibinfo{author}{Kim, D.},
  \bibinfo{author}{Alfi, D.M.}, \bibinfo{author}{Yap, P.T.},
  \bibinfo{author}{Xia, J.J.}, \bibinfo{author}{Shen, D.},
  \bibinfo{year}{2019}.
\newblock \bibinfo{title}{Estimating reference bony shape model for
  personalized surgical reconstruction of posttraumatic facial defects}, in:
  \bibinfo{editor}{Shen, D.}, \bibinfo{editor}{Liu, T.},
  \bibinfo{editor}{Peters, T.M.}, \bibinfo{editor}{Staib, L.H.},
  \bibinfo{editor}{Essert, C.}, \bibinfo{editor}{Zhou, S.},
  \bibinfo{editor}{Yap, P.T.}, \bibinfo{editor}{Khan, A.} (Eds.),
  \bibinfo{booktitle}{Medical Image Computing and Computer Assisted
  Intervention -- MICCAI 2019}, \bibinfo{publisher}{Springer International
  Publishing}, \bibinfo{address}{Cham}. pp. \bibinfo{pages}{327--335}.
\bibitem[{Yang et~al.(2023a)Yang, Zhang, Chen, Xu, Sun and Ma}]{YangQiMPED}
\bibinfo{author}{Yang, Q.}, \bibinfo{author}{Zhang, Y.}, \bibinfo{author}{Chen,
  S.}, \bibinfo{author}{Xu, Y.}, \bibinfo{author}{Sun, J.},
  \bibinfo{author}{Ma, Z.}, \bibinfo{year}{2023}a.
\newblock \bibinfo{title}{Mped: Quantifying point cloud distortion based on
  multiscale potential energy discrepancy}.
\newblock \bibinfo{journal}{IEEE Transactions on Pattern Analysis and Machine
  Intelligence} \bibinfo{volume}{45}, \bibinfo{pages}{6037--6054}.
\newblock \DOIprefix\doi{10.1109/TPAMI.2022.3213831}.
\bibitem[{Yang et~al.(2023b)Yang, Guo, Xiong, Liu, Pan, Wang, Tong and
  Guo}]{yangyuswin}
\bibinfo{author}{Yang, Y.Q.}, \bibinfo{author}{Guo, Y.X.},
  \bibinfo{author}{Xiong, J.Y.}, \bibinfo{author}{Liu, Y.},
  \bibinfo{author}{Pan, H.}, \bibinfo{author}{Wang, P.S.},
  \bibinfo{author}{Tong, X.}, \bibinfo{author}{Guo, B.}, \bibinfo{year}{2023}b.
\newblock \bibinfo{title}{Swin3d: A pretrained transformer backbone for 3d
  indoor scene understanding}.
\newblock \bibinfo{journal}{arXiv preprint arXiv:2304.06906} .
\bibitem[{Yin et~al.(2018)Yin, Huang, Cohen-Or and Zhang}]{p2p}
\bibinfo{author}{Yin, K.}, \bibinfo{author}{Huang, H.},
  \bibinfo{author}{Cohen-Or, D.}, \bibinfo{author}{Zhang, H.},
  \bibinfo{year}{2018}.
\newblock \bibinfo{title}{P2p-net: bidirectional point displacement net for
  shape transform}.
\newblock \bibinfo{journal}{ACM Trans. Graph.} \bibinfo{volume}{37}.
\newblock \URLprefix \url{https://doi.org/10.1145/3197517.3201288},
  \DOIprefix\doi{10.1145/3197517.3201288}.
\bibitem[{Yu et~al.(2023)Yu, Rao, Wang, Lu and Zhou}]{YuXumin}
\bibinfo{author}{Yu, X.}, \bibinfo{author}{Rao, Y.}, \bibinfo{author}{Wang,
  Z.}, \bibinfo{author}{Lu, J.}, \bibinfo{author}{Zhou, J.},
  \bibinfo{year}{2023}.
\newblock \bibinfo{title}{Adapointr: Diverse point cloud completion with
  adaptive geometry-aware transformers}.
\newblock \bibinfo{journal}{IEEE Transactions on Pattern Analysis and Machine
  Intelligence} \bibinfo{volume}{45}, \bibinfo{pages}{14114--14130}.
\newblock \DOIprefix\doi{10.1109/TPAMI.2023.3309253}.
\bibitem[{Zammit et~al.(2023)Zammit, Ettinger, Sanati-Mehrizy and
  Susarla}]{Zammit}
\bibinfo{author}{Zammit, D.}, \bibinfo{author}{Ettinger, R.E.},
  \bibinfo{author}{Sanati-Mehrizy, P.}, \bibinfo{author}{Susarla, S.M.},
  \bibinfo{year}{2023}.
\newblock \bibinfo{title}{Current trends in orthognathic surgery}.
\newblock \bibinfo{journal}{Medicina} \bibinfo{volume}{59}.
\newblock \URLprefix \url{https://www.mdpi.com/1648-9144/59/12/2100},
  \DOIprefix\doi{10.3390/medicina59122100}.
\bibitem[{Zhang et~al.(2024a)Zhang, Liu, Wu and Wang}]{ZhangDan}
\bibinfo{author}{Zhang, D.}, \bibinfo{author}{Liu, N.}, \bibinfo{author}{Wu,
  Z.}, \bibinfo{author}{Wang, X.}, \bibinfo{year}{2024}a.
\newblock \bibinfo{title}{3d craniofacial similarity calculation and
  craniofacial relationships analysis based on spectral analysis method}.
\newblock \bibinfo{journal}{Multimedia Tools and Applications}
  \bibinfo{volume}{83}, \bibinfo{pages}{14063--14084}.
\bibitem[{Zhang et~al.(2024b)Zhang, Jie, He, Zhu, Xie, Liu, Mo and
  Wang}]{zhangrunshi}
\bibinfo{author}{Zhang, R.}, \bibinfo{author}{Jie, B.}, \bibinfo{author}{He,
  Y.}, \bibinfo{author}{Zhu, L.}, \bibinfo{author}{Xie, Z.},
  \bibinfo{author}{Liu, Z.}, \bibinfo{author}{Mo, H.}, \bibinfo{author}{Wang,
  J.}, \bibinfo{year}{2024}b.
\newblock \bibinfo{title}{Craniomaxillofacial bone segmentation and landmark
  detection using semantic segmentation networks and an unbiased heatmap}.
\newblock \bibinfo{journal}{IEEE Journal of Biomedical and Health Informatics}
  \bibinfo{volume}{28}, \bibinfo{pages}{427--437}.
\newblock \DOIprefix\doi{10.1109/JBHI.2023.3337546}.
\bibitem[{Zhang et~al.(2024c)Zhang, Mo, Hu, Jie, Xu, He, Ke and
  Wang}]{zhanglandmark}
\bibinfo{author}{Zhang, R.}, \bibinfo{author}{Mo, H.}, \bibinfo{author}{Hu,
  W.}, \bibinfo{author}{Jie, B.}, \bibinfo{author}{Xu, L.},
  \bibinfo{author}{He, Y.}, \bibinfo{author}{Ke, J.}, \bibinfo{author}{Wang,
  J.}, \bibinfo{year}{2024}c.
\newblock \bibinfo{title}{Super-resolution landmark detection networks for
  medical images}.
\newblock \bibinfo{journal}{Computers in Biology and Medicine}
  \bibinfo{volume}{182}, \bibinfo{pages}{109095}.
\newblock \URLprefix
  \url{https://www.sciencedirect.com/science/article/pii/S0010482524011806},
  \DOIprefix\doi{https://doi.org/10.1016/j.compbiomed.2024.109095}.
\bibitem[{Zhang et~al.(2024d)Zhang, Mo, Wang, Jie, He, Jin and
  Zhu}]{zhang2024utsrmorph}
\bibinfo{author}{Zhang, R.}, \bibinfo{author}{Mo, H.}, \bibinfo{author}{Wang,
  J.}, \bibinfo{author}{Jie, B.}, \bibinfo{author}{He, Y.},
  \bibinfo{author}{Jin, N.}, \bibinfo{author}{Zhu, L.}, \bibinfo{year}{2024}d.
\newblock \bibinfo{title}{Utsrmorph: A unified transformer and superresolution
  network for unsupervised medical image registration}.
\newblock \bibinfo{journal}{IEEE Transactions on Medical Imaging} .
\bibitem[{Zhao et~al.(2019)Zhao, Jiang, Fu and Jia}]{ZhaoHengshuangweb}
\bibinfo{author}{Zhao, H.}, \bibinfo{author}{Jiang, L.}, \bibinfo{author}{Fu,
  C.W.}, \bibinfo{author}{Jia, J.}, \bibinfo{year}{2019}.
\newblock \bibinfo{title}{Pointweb: Enhancing local neighborhood features for
  point cloud processing}, in: \bibinfo{booktitle}{Proceedings of the IEEE/CVF
  conference on computer vision and pattern recognition}, pp.
  \bibinfo{pages}{5565--5573}.
\bibitem[{Zhao et~al.(2021)Zhao, Jiang, Jia, Torr and Koltun}]{ZhaoHengshuang}
\bibinfo{author}{Zhao, H.}, \bibinfo{author}{Jiang, L.}, \bibinfo{author}{Jia,
  J.}, \bibinfo{author}{Torr, P.}, \bibinfo{author}{Koltun, V.},
  \bibinfo{year}{2021}.
\newblock \bibinfo{title}{Point transformer}, in: \bibinfo{booktitle}{2021
  IEEE/CVF International Conference on Computer Vision (ICCV)}, pp.
  \bibinfo{pages}{16239--16248}.
\newblock \DOIprefix\doi{10.1109/ICCV48922.2021.01595}.

\end{thebibliography}

\end{document}